%% file: main.tex
\documentclass{article} 
\usepackage[final]{template}

\input{math_commands.tex}

\usepackage{hyperref}
\usepackage{url}
\usepackage{graphicx}

\usepackage{titletoc}
\usepackage{booktabs}
\usepackage{multirow}
\usepackage{tabularx}
\usepackage{graphicx} 
\usepackage[export]{adjustbox} 
\usepackage{makecell}   
\usepackage{array} 
\usepackage{pgfplots}
\usepackage{subcaption}
\usepackage{wrapfig}
\usepackage{tcolorbox}
\tcbuselibrary{breakable}
\usepackage{enumitem}
\usepackage{marvosym}
\usepackage{listings}
\usepackage{xspace}
\tcbuselibrary{skins}

\usepackage{xcolor}
\definecolor{citecolor}{HTML}{0071BC}
\definecolor{linkcolor}{HTML}{1A73E8}
\definecolor{urlcolor}{HTML}{0B8043}

\usepackage{hyperref}
\hypersetup{
  colorlinks=true,
  citecolor=citecolor,
  pagebackref=false,
  breaklinks=true,
  bookmarks=false
}
\tcbset{
  simpleprompt/.style={
    breakable, enhanced,
    boxrule=0.7pt, arc=2mm,
    colframe=black!80,              
    left=6pt,right=6pt,top=6pt,bottom=6pt,
    fonttitle=\bfseries,
  }
}
\newtcolorbox{prompt}[2][]{simpleprompt,
  colback=citecolor!6!white,        
  title={#2}, #1}
  
\pgfplotsset{compat=1.18}

\newcommand{\bench}{AUI-Gym}

\newcommand{\ie}{\textit{i.e.,}}
\newcommand{\eg}{\textit{e.g.,}}

\definecolor{darkred}{HTML}{8B0000}
\definecolor{AuiTask}{HTML}{3966BF} 
\definecolor{AuiRule}{HTML}{00897B} 

\tcbset{
  aibox/.style={
    width=\linewidth,
    top=8pt,
    bottom=4pt,
    colback=citecolor!5!white,
    colframe=black,
    colbacktitle=black,
    enhanced,
    center,
    attach boxed title to top left={yshift=-0.1in,xshift=0.15in},
    boxed title style={boxrule=0pt,colframe=white,},
  }
}
\newtcolorbox{AIbox}[2][]{aibox,title=#2,#1}

\newcommand{\githubemoji}{%
  \raisebox{-0.20\height}{%
    \hspace{0.15em}
    \includegraphics[height=0.9em]{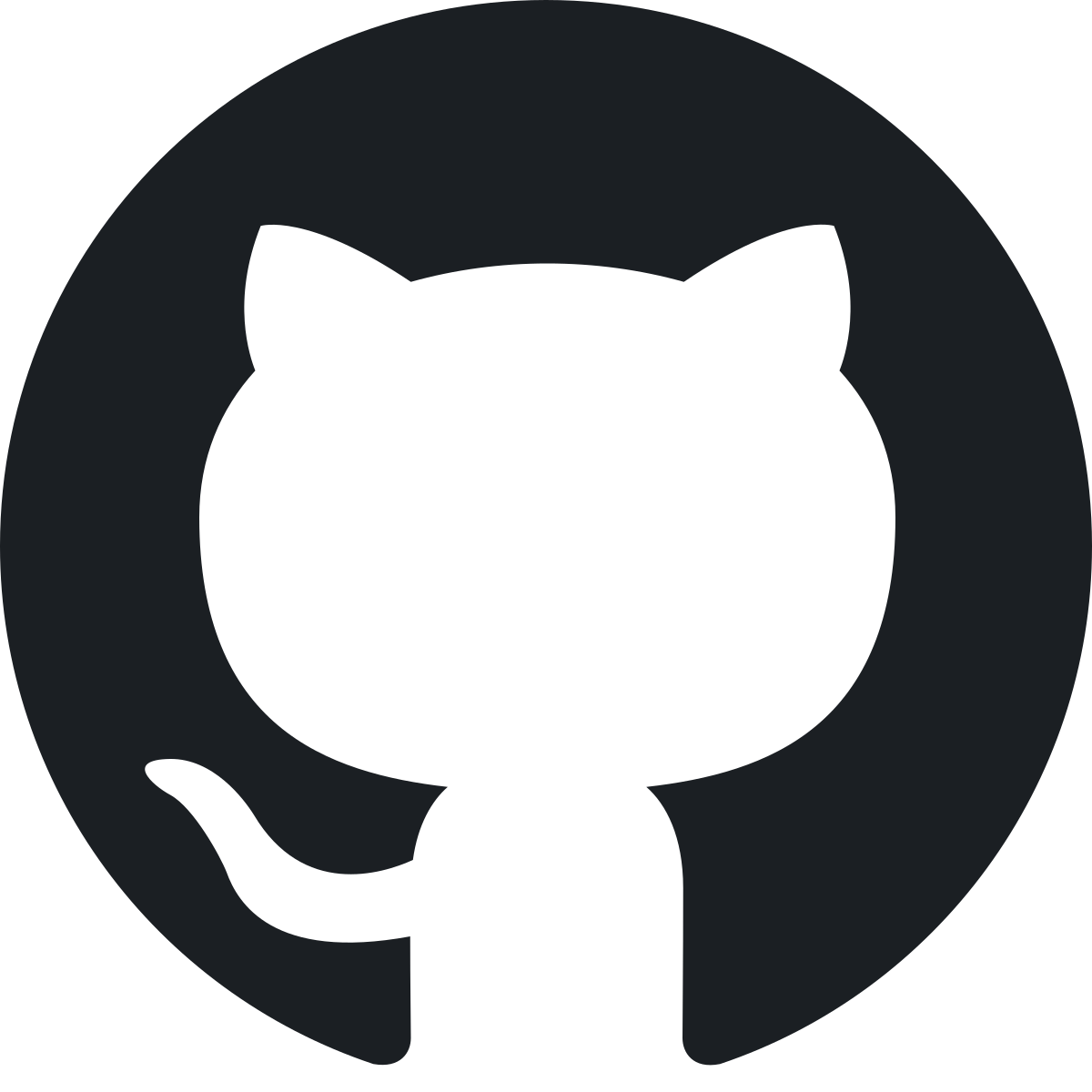}%
  }\xspace
}
\newcommand{\houseemoji}{%
  \raisebox{-0.20\height}{%
    \hspace{0.15em}
    \includegraphics[height=0.9em]{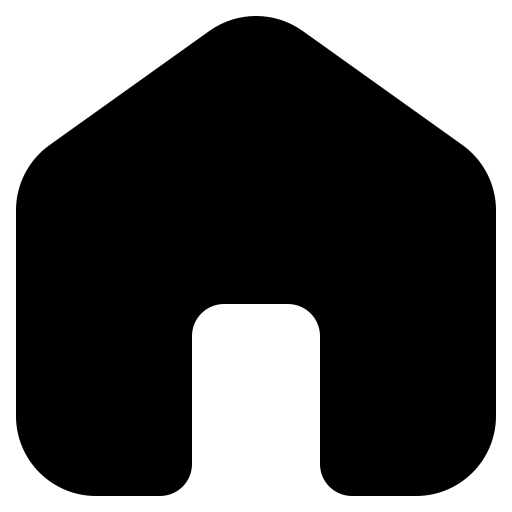}%
  }\xspace
}
\newcommand{\hfemoji}{%
  \raisebox{-0.20\height}{%
    \hspace{0.1em}
    \includegraphics[height=1em]{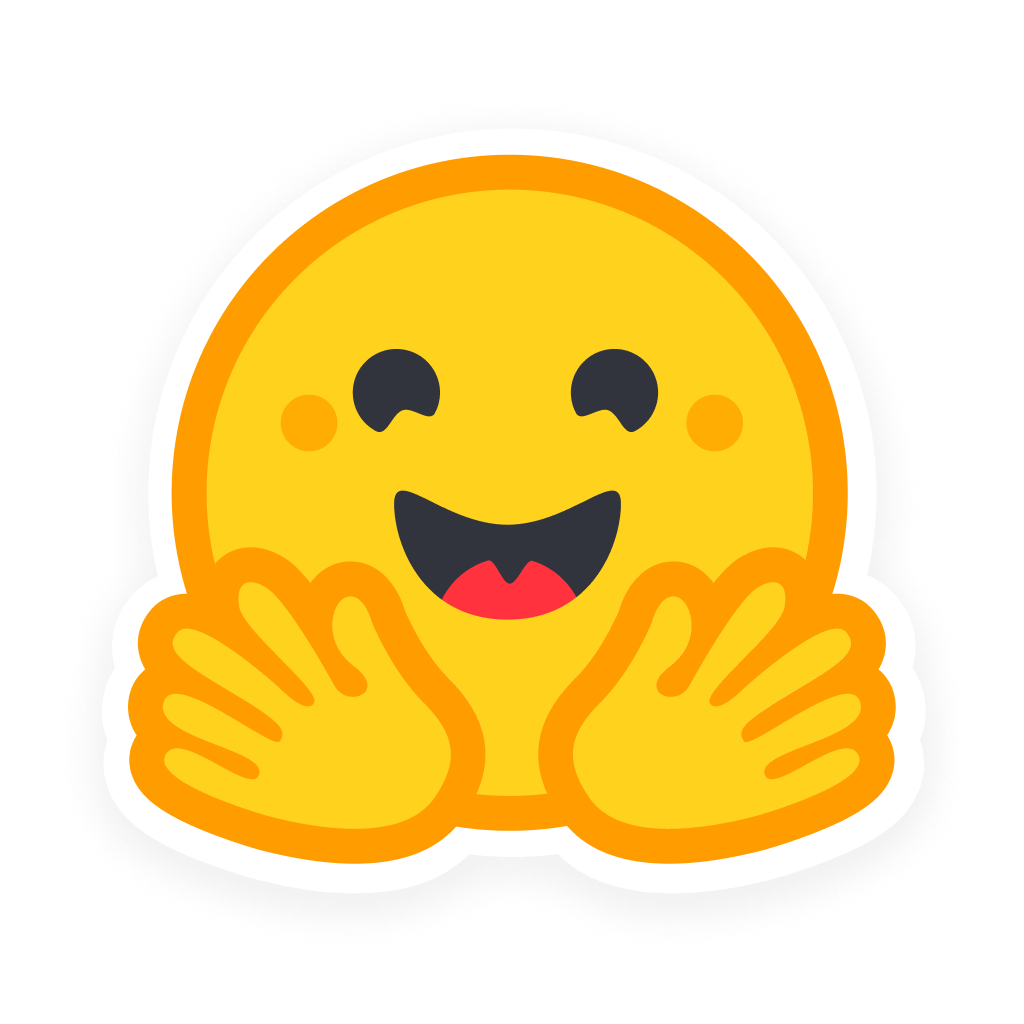}%
  }\xspace
}

\title{Computer-Use Agents as Judges for\\Generative User Interface}

\author{%
$^1$Kevin Qinghong Lin\thanks{Equal contribution. \Letter\ Corresponding author}\quad
$^2$Siyuan Hu\footnotemark[1]\quad
$^3$Linjie Li\quad
$^3$Zhengyuan Yang\\
\textbf{$^3$Lijuan Wang\quad
$^1$Philip Torr\quad
$^2$Mike Zheng Shou\textsuperscript{\Letter}}
\\[4px]
$^1$University of Oxford\quad
$^2$Show Lab, National University of Singapore\quad
$^3$Microsoft \\
[2mm]
\houseemoji~{Homepage:}~\href{https://showlab.github.io/AUI}{\texttt{https://showlab.github.io/AUI}}\\
\githubemoji~{Code: }~\href{https://github.com/showlab/AUI}{\texttt{https://github.com/showlab/AUI}}\\
\hfemoji~{Demo:}~\href{https://huggingface.co/spaces/showlab/AUI}{\texttt{https://huggingface.co/spaces/showlab/AUI}}\\
}

\makeatletter
\newcommand{\@trackname}{}
\makeatother

\begin{document}

\maketitle

\begin{abstract}
Computer-Use Agents (CUA) are becoming increasingly capable of autonomously operating digital environments through Graphical User Interfaces (GUI). Yet, most GUI remain designed primarily for humans—prioritizing aesthetics and usability—forcing agents to adopt human-oriented behaviors that are unnecessary for efficient task execution.
At the same time, rapid advances in coding-oriented language models (Coder) have transformed automatic GUI design. This raises a fundamental question: \textit{Can CUA as judges to assist Coder for automatic GUI design}?
To investigate, we introduce \textbf{\bench}, a benchmark for {A}utomatic G{UI} development spanning 52 applications across diverse domains. Using  language models, we synthesize 1560 tasks that simulate real-world scenarios. To ensure task reliability, we further develop a verifier that programmatically checks whether each task is executable within its environment.
Building on this, we propose a \textbf{Coder-CUA in Collaboration} framework: the Coder acts as Designer, generating and revising websites, while the CUA serves as Judge, evaluating functionality and refining designs. Success is measured not by visual appearance, but by task solvability and CUA navigation success rate. 
To turn CUA feedback into usable guidance, we design a \textbf{CUA Dashboard} that compresses multi-step navigation histories into concise visual summaries, offering interpretable guidance for iterative redesign.
By positioning agents as both designers and judges, our framework shifts interface design toward agent-native efficiency and reliability. 
Our work takes a step toward shifting agents from passive use toward active participation in digital environments. 
Our code and dataset are available at \url{https://github.com/showlab/AUI}.
\end{abstract}

\input{section/intro}
\input{section/relatedwork}
\input{section/bench}
\input{section/method}
\input{section/exps}
\input{section/conclusion}

\bibliography{main}
\bibliographystyle{unsrt}
\input{section/appendix}

\end{document}

%% file: math_commands.tex
\usepackage{amsmath,amsfonts,bm}

\def\eqref#1{equation~\ref{#1}}

\def\1{\bm{1}}

\DeclareMathAlphabet{\mathsfit}{\encodingdefault}{\sfdefault}{m}{sl}
\SetMathAlphabet{\mathsfit}{bold}{\encodingdefault}{\sfdefault}{bx}{n}



%% file: section/intro.tex
\section{Introduction}
Recent advances in language agents have shown that \textbf{Computer-Use Agents}~\cite{openaioperator, claude} can autonomously operate within GUIs—performing tasks such as online shopping by sequentially clicking through multiple buttons~\cite{webarena}.
However, today’s environments remain fundamentally human-centric, optimized for aesthetics and usability through features like dynamic animations or colorful layouts. To adapt to these settings, researchers typically train CUA on large-scale human demonstration trajectories, click logs, or static screenshots~\cite{aguvis, showui, uitars}, effectively forcing agents to imitate human behavior. This approach binds automation to human-oriented design choices, where stylistic details crucial for humans are redundant for agents whose primary objective is efficient task completion.
In parallel, coding-oriented language models—\textbf{Coders}—have already demonstrated strong capabilities, capable of generating functional HTML pages or even entire websites from a single instruction~\cite{design2code}. Yet these outputs remain confined to {human-facing loops}: even when generated by agents, interfaces are still optimized for human use rather than agent-native interaction.

Both CUA and Coders thus exhibit remarkable potential for automation and design. This motivates a fundamental question:
Can CUA assists Coders redesign UIs in an automatic manner—where environments are created for, and evaluated by, agents themselves, with CUA acting as judges?
In this work, we reconceptualize the UI as a tunable environment. The core idea is to employ the \textbf{\textit{Coder as Designer}}—responsible for initializing and revising UIs—while the \textbf{\textit{CUA acts as Judges}}, navigating through tasks and collecting interaction trajectories as feedback.

\begin{figure}[!t]
    \centering
    \includegraphics[width=1\textwidth]{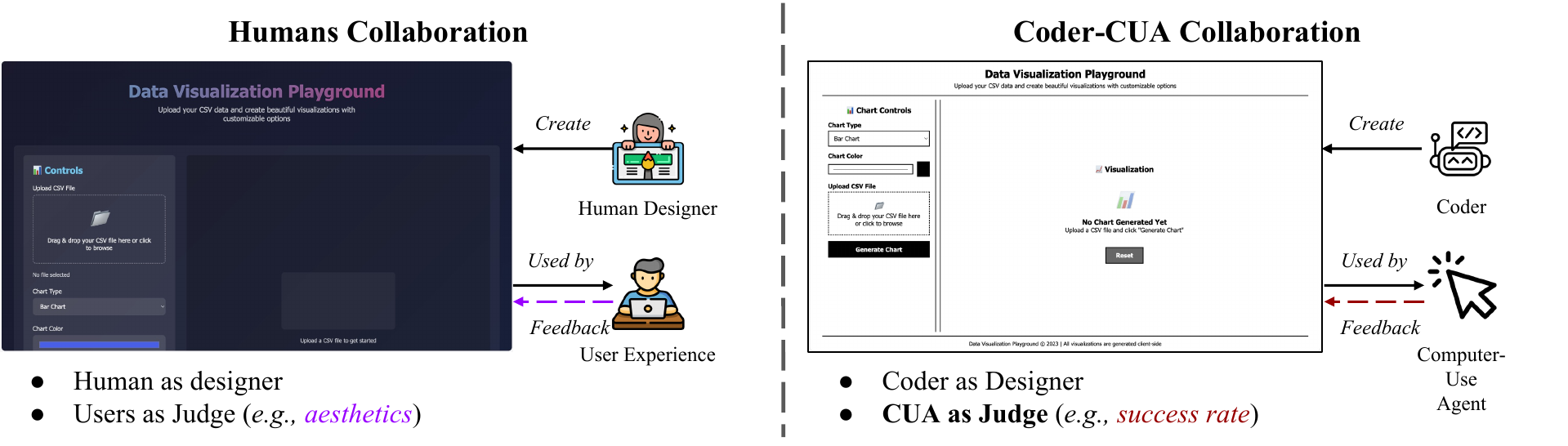}
\caption{
\textbf{Illustration of Humans Collaboration \textit{vs.} our Coder-CUA Collaboration in term of UI designs.} 
\textbf{Left:} Most GUIs are designed by humans and optimized for user experience (\eg~aesthetics), forcing trained agents to adapt to human-oriented behaviors. 
\textbf{Right:} Our Coder-CUA Collaboration framework leverages Coder as Designer and CUA as Judge together, enabling more reliable task execution and improved usability for agents. 
}
\label{fig:teaser}
\end{figure}

As no existing testbed aligns with our goal, we introduce \bench~to pioneer evaluation in this setting. \bench~automatically develops websites across 52 applications spanning six domains (apps, landing pages, games, interactive demos, tools, and utilities). Unlike most coders that focus on single-page generation, \bench~requires agents to produce fully automated, executable, application-level designs with an emphasis on functional completeness.
Enabling sufficient, scalable, and human-free evaluation is non-trivial. To simulate realistic usage scenarios, we prompt GPT-5 to propose 30 candidate tasks per application, yielding 1560 tasks in total. These tasks are then validated by humans. To ensure that each website can be reliably tested, GPT-5 also generates a customized rule-based functional checker for individual task, determining whether the task is feasible within the given interface.
This infrastructure establishes a human-free, reliable foundation for subsequent CUA exploration and feedback-driven UI refinement.

To this end, we develop a \textbf{Coder--CUA collaboration} framework. The \textit{Coder acts as Designer}, responsible for UI initialization and refinement, while the \textit{CUA serves as Judge}, supplying feedback. The central challenge is how to transform raw CUA interactions into effective revision signals from an agent perspective. We address this through two complementary dimensions of feedback: \textbf{(a) CUA Navigation}, where the agent executes tasks through atomic actions such as clicks or typing and judges success or failure; and \textbf{(b) Task Solvability}, where unsolvable tasks are accumulated as functionality failures and returned to the Coder as precise indicators of missing features.  
CUA navigation produces long, multi-step trajectories interleaved with screenshots, making direct feedback difficult to interpret. To overcome this, we introduce the \textbf{CUA Dashboard}, which condenses each task, its outcome, actions, and intermediate states into a single $1920\times1080$ image. Rather than storing every screenshot, the dashboard highlights only key interactive regions, with region sizes adaptively scaled by the number of steps. This dynamic design reduces redundancy by average 76.2\% while preserving essential cues, offering a clear step-by-step view of how the CUA perceives and acts on the interface. As a result, success and failure points become immediately visible, and the dashboard provides concise, interpretable feedback that the Coder to guide iterative UI redesign.

Our empirical results show that while state-of-the-art Coders can generate complete GUIs that appear suitable to humans, they still encounter notable limitations:  
\textbf{(i) Task solvability as a foundation.} Initial UIs often fail to capture many practical scenarios, resulting in low usability. However, by collecting failure cases, the Coder can readily boost performance through language-based functional summarization.
\textbf{(ii) CUA navigation as a key bottleneck.} Even when UIs achieve high functional completeness, CUAs initially exhibit low success rates due to the complexity of multi-step navigation. Through our Coder--CUA collaboration, we substantially improve navigation success rates, particularly showing that CUA feedback-driven redesigns---such as \textit{de-stylization}, increased contrast, and simplified layouts---significantly enhance CUA execution.
Together, these findings highlight the promising potential of agents for automatic UI design and testing, improving both task success and robustness.
To summarize, our contributions are threefold:

\begin{enumerate}[leftmargin=*, itemsep=0pt, topsep=0pt]
\item \textbf{\bench: a scalable testbed for automatic GUI development and testing}, covering 52 applications across six domains with 1560 GPT-5–proposed, human-validated tasks and per-task rule-based checkers. This enables human-free development of automatic UI creation and testing.

\item \textbf{Coder--CUA framework with CUA Dashboard.} The {Coder} initializes and refines UIs while the {CUA} judges via two signals: navigation outcomes and task solvability. 
A single-image $1K$ {CUA Dashboard} compresses task goal, actions, intermediate states, and outcome by highlighting key interactive regions with adaptive scaling, reducing visual tokens by {76.2\% on average} while preserving essential cues for redesign.

\item \textbf{Evaluation Insights: What kind of UI do Agents prefer?} Task solvability is foundational yet readily improved via failure-driven functional summarization, whereas CUA navigation is the main bottleneck. Feedback-driven redesigns (\eg~de-stylization, higher contrast, simplified layouts) substantially raise execution success and overall robustness.
\end{enumerate}

%% file: section/relatedwork.tex
\section{Related Works}
\subsection{Computer-Use Agents}
Recent studies reveal the potential of LLMs beyond language modeling, with advancements in demonstrating their ability to autonomously complete complex tasks using tool integration~\citep{toolformer} like humans. This has prompted the development of GUI automation agents that learn to operate digital user interfaces by imitating human trajectories. This learning is primarily achieved in two ways: 
(i) by steering general multimodal foundation models with in-context human trajectory examples, and the general models perceive the UI through intermediate representations like HTML, accessibility trees~\citep{workarena, seeact}, Optical Character Recognition~\citep{omniparser}, or Set of Masks~\citep{setofmask}. 
(ii) by pre-training specialized GUI foundation models through extensive supervised fine-tuning or reinforcement learning on large-scale vision-text UI data  (\eg~screenshots and instructions)~\citep{showui, uground, videogui, uivision}. While foundational, these data-driven approaches suffered from heavy requirements for high-quality human trajectories to achieve agent performance improvements. 
Despite their methodological differences, these approaches share a common, agent-centric paradigm, focusing on improving the agent's capabilities to navigate a static and often complex environment. Notably, we investigate a complementary approach. Instead of adapting the agent, we explore how to dynamically tune the environment to enhance the performance of a frozen agent.

\subsection{Automatic Software Designs}
Besides CUAs, there have been extensive research on software automation, automatic interface design~\citep{uilayout, adaptiveUIgrammar} and generation~\citep{design2code, pix2code, unlocking}. Programmatic and semantic UI components—such as accessibility layers, ARIA tags, and declarative interface frameworks (\eg~React Native, Flutter)—illustrate how environments can be annotated or abstracted for automated processes. Similarly, benchmarks in automated software interaction, such as WebArena~\citep{webarena} and GAIA~\citep{gaia}, assume agent operates within fixed, human-oriented systems for task automation. More recently, embodied AI environments (\eg~ALFRED~\citep{alfred}, Habitat~\citep{habitat}, MineDojo~\citep{minedojo}) show how environments can be crafted to accelerate agent training, though primarily in physical or simulated domains. These efforts highlight the growing recognition that environments themselves can be reimagined for machine interaction, yet a systematic framework for designing agent-centric digital environments in everyday computing remains absent.

%% file: section/bench.tex
\section{\bench~Benchmark}

\subsection{Task Definitions}
We develop \bench~for automatic GUI development and testing.  
Given a language user query $\mathcal{Q}$ as input and several available agents (\eg~Coder or CUA), the output is a complete website that serves as a \textit{tunable} environment $\mathcal{E}$.
We detail the input and output respectively below.

\begin{wrapfigure}{r}{0.6\linewidth}
    \centering
    \vspace{-1em}
    \includegraphics[width=\linewidth]{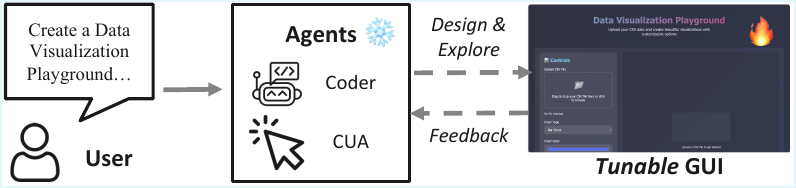} 
    \caption{\textbf{\bench~task definition.} A user issues a request (\eg~“Create a Data Visualization Playground”), and agents (\eg~Coder or CUA) interact with the GUI through design, exploration, and feedback. In this setup, the GUI serves as a \textit{tunable} environment.}
    \label{fig: motivation}
    \vspace{-2em}
\end{wrapfigure}

\textbf{Input Query $\mathcal{Q}$.}
Since the outcome is a website, the user query $\mathcal{Q}$ should be both descriptive and concrete. To this end, we explicitly standardize queries into the structured format illustrated above. This supplements the query with a name, goal, functional features, and UI theme.

\textbf{Output website $\mathcal{E}$.}
The website is an application-level deliverable that must be fully functional, going beyond a static page to support navigation, transitions, button interactions, and completion of functional goals, with the objective of maximizing the agent's success rate.  
Constructing an effective evaluation framework in this setting is non-trivial and introduces several challenges.
We next present our scalable, automatic solutions.

\begin{prompt}{Input formulation}
Create a single-page app in a single HTML file with the following requirements: \\ 
- Name: \{Camping Gear Checklist\} \\ 
- Goal: \{Track gear for camping trips\}. \\ 
- Features: \{Checklist items, weight calculator, save lists.\} \\ 
- Theme: \{The UI should be outdoor-themed.\}
\label{prompt}
\end{prompt}

\vspace{-1em}
\subsection{Task Creation}
\begin{figure}[!t]
    \centering
    \includegraphics[width=1\textwidth]{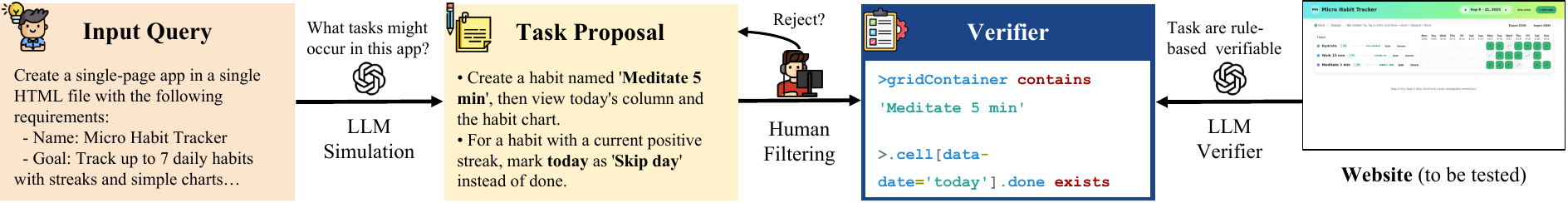}
\caption{\textbf{\bench~construction pipeline.} 
\textbf{(i)} An input query specifies the app requirements. 
\textbf{(ii)} GPT-5 proposes candidate tasks with explicit goals. 
\textbf{(iii)} Humans filter and refine tasks using domain-specific principles. 
\textbf{(iv)} A test-time Verifier reads the website HTML and generates task-specific, rule-based checkers to validate success on the to-be-tested website.}
\label{fig:curation}
\end{figure}

The full curation pipeline is illustrated in Fig.\ref{fig:curation}.
To construct the benchmark, we collect 52 task prompts from OpenAI's playground~\footnote{https://github.com/openai/gpt-5-coding-examples}, covering multiple domains. 

\textbf{Synthesize candidate tasks $\mathcal{T}$.}
Applications are typically designed to support a variety of relevant tasks, and a key evaluation is whether they can smoothly handle such tasks.  
We leverage GPT-5~\cite{gpt5blog} to synthesize diverse user requirements: given an instruction $\mathcal{I}$, it generates a set of candidate tasks $\mathcal{T}$ that simulate practical usage.  
As illustrated in Fig.~\ref{fig:curation}, for the application `\textit{Micro Habit Tracker}', an example task is: ``\textcolor{AuiTask}{\textit{Create a habit named `Meditate 5 min,' then view today's column and the habit chart.}}''  
These tasks serve as fine-grained probes that capture the potential demands of the environment $\mathcal{E}$.

\textbf{Manual quality control.}
As the tasks are automatically generated by GPT-5, human oversight is required to ensure their quality.  
Different applications demand different characteristics: for example, tasks for game UIs should emphasize interactivity and control, while tasks for utility tools should capture information accessibility and workflow patterns.  
To this end, humans define domain-specific principles and filter out low-quality tasks (\eg~trivial clicks) or nonsensical ones (\eg~beyond the application scope), ambiguous queries (cross-application), ensuring that the proposed tasks remain concrete, meaningful and aligned with each domain’s design philosophy.

\textbf{Data Statistics.}
\input{tex/stats_subset}
Based on the above strategy, we obtain 30 tasks for each application.  
The benchmark spans 52 web applications across six domains, yielding a total of 1,560 tasks and enabling comprehensive evaluation across diverse applications.  
As illustrated in Table~\ref{tab:stats}, the domains include:
{(i) App}, general-purpose applications; 
{(ii) Landing}, commercial and promotional interfaces; 
{(iii) Game}, puzzle and arcade-style challenges; 
{(iv) Interactive}, dynamic user engagement with real-time feedback; 
{(v) Tool}, specialized utilities; and 
{(vi) Utility}, everyday organizational support. 
This diverse coverage captures distinct GUI challenges-—ensuring robust evaluation across varied interaction paradigms and functional complexities.

\subsection{Evaluation with Verifiers}
\label{sec:verifier}
Even with the proposed tasks, it remains challenging to determine whether a given GUI can truly satisfy them, as websites are interactive and highly diverse environments.  
More importantly, since the \textit{GUIs are generated at test time}, it is difficult to design fixed standards that generalize across all cases, given the variety of possible implementation approaches.
A naive solution is to adopt a VLM-as-Judge approach, but this inevitably introduces bias and uncertainty. Ideally, the most reliable solution would be concrete functional checks with manual validation, yet this approach is prohibitively expensive and labor-intensive.

\begin{wrapfigure}{r}{0.48\textwidth}
\vspace{-2em}
\begin{lstlisting}[language=Python,basicstyle=\ttfamily\scriptsize]
Verifier(input = GUI_HTML, task):
    analyze elements and states
    if task solvable:
        return (Yes, function_checker)
    else:
        return (No, None)
\end{lstlisting}
\vspace{-2em}
\end{wrapfigure}

To address this, we define a \textbf{Verifier} ${\mathcal{V}(\cdot)}$ powered by GPT-5 at test time, which takes as input a candidate GUI together with a specific task.  
It analyzes the available elements and states, reasoning over the presence of required UI components, their properties, and potential interaction paths.  
If the task is deemed solvable, the Verifier produces a task-specific verification \textbf{function checker} $\widetilde{\mathcal{V}}(\cdot)$ (by \texttt{JavaScript}) that encodes the success condition by element status; otherwise, the task is discarded as invalid, preventing noisy or unachievable goals from disrupting evaluation.
Such as in Fig.\ref{fig:curation}, for task ``\textcolor{AuiTask}{\textit{Create a habit named `Meditate 5 min,' then view today's column and the habit chart.}}'', based on the candidate website (right), the verifier generates the rule \textcolor{AuiRule}{\texttt{gridContainer contains 'Meditate 5min'}}
In this way, the Verifier is customized for each website and each task at test time, ensuring reliable validation.

\textbf{Metrics.} 
With the support of function checkers as reliable verification, we can ensure that a website is both actionable and workable for the CUA.  
This further allows us to evaluate whether tasks are completed after CUA navigation, thereby measuring task success rate within the UI environment.
In this way, we devise the following measure:

\textbf{(i) CUA Success Rate (SR).}  
This measures the average success rate over all tasks executed by CUA.  
If CUA successfully completes a task, it is counted as a success; otherwise, it is counted as a failure.  
Notably, if the Coder fails to yield a functional checker, the task is counted as a failure.  
\begin{equation}
    \text{SR} = \frac{1}{|\mathcal{T}|} \sum_{t \in \mathcal{T}} \mathbf{1}\left(\text{task $t$ is successfully completed}\right),
    \label{eq:sr}
\end{equation}
where $\mathcal{T}$ denotes the set of all tasks and $\mathbf{1}\{\cdot\}$ is the indicator function.

\textbf{(ii) Function Completeness (FC).}  
While CUA performance reflects the ultimate goal, it may be sparse if most CUAs fail to complete tasks.  
Therefore, we devise a second metric to evaluate only whether the Coder-created website functionally supports the task (valid), independent of CUA navigation.  
This metric reflects task validity and serves as a more basic measure.  
\begin{equation}
    \text{FC} = \frac{1}{|\mathcal{T}|} \sum_{t \in \mathcal{T}} \mathbf{1}\{\text{a functional checker exists for task $t$}\}.
    \label{eq:fc}
\end{equation}

%% file: tex/stats_subset.tex
\begin{table}[!t]
  \centering
  \scriptsize
  \vspace{-2em}
\caption{\textbf{Examples of App domains in \bench.} For each domain, we show a website created by GPT-5, paired with 30 tasks (\textcolor{AuiTask}{blue}) simulating real-world usage. Each task is further linked to a rule-based verifier (\textcolor{AuiRule}{green}). See full distribution and examples in Tab.\ref{tab:stats:full}.}
  \label{tab:stats}
  \setlength{\tabcolsep}{4pt}
  \renewcommand{\arraystretch}{1.12}
  \begingroup
  \begin{tabularx}{\textwidth}{lccXc}
    \toprule
    \textbf{Domain} & \textbf{\#Apps} & \makecell{\textbf{Percen-}\\\textbf{tage}} & \textbf{Example Instruction} & \textbf{GUI created by GPT-5} \\
    \midrule
    App & \makecell[t]{11} & \makecell[t]{21\%} &
    \parbox[t]{\linewidth}{\raggedright\scriptsize
      Create a single-page app in a single HTML file with the following requirements:\newline
      - Name: Healthy Meal Tracker\newline
      - Goal: Log meals and nutrition info.\newline
      - Features: Ingredient list, calories per meal, daily summary.\newline
      - The UI should be clean with food icons.\newline
      \textcolor{AuiTask}{\textbf{Task: Add five meals for today's date (any names/ingredients) so today's meal count reaches at least 5.}}\newline
      \textcolor{AuiRule}{\textbf{Rule: \texttt{\#dailyMealCount >= 5}}}
    } &
    \includegraphics[width=3.5cm, valign=t]{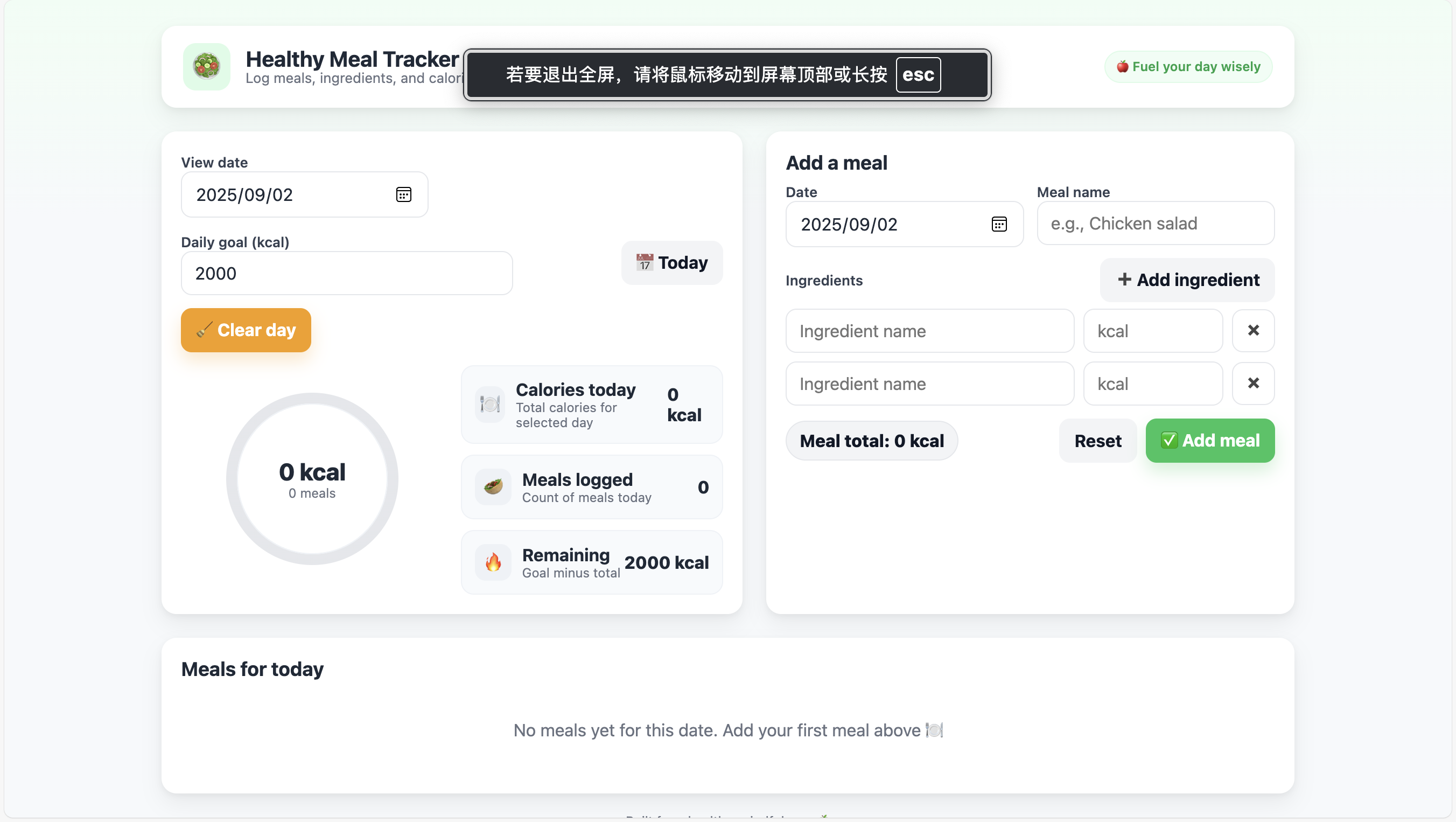} \\
    \bottomrule
  \end{tabularx}
  \vspace{-2em}
  \endgroup
\end{table}

%% file: section/method.tex
\vspace{-1.5em}

\section{CUA–Coder in Collaboration}
\begin{figure}[!t]
    \centering
    \includegraphics[width=\textwidth]{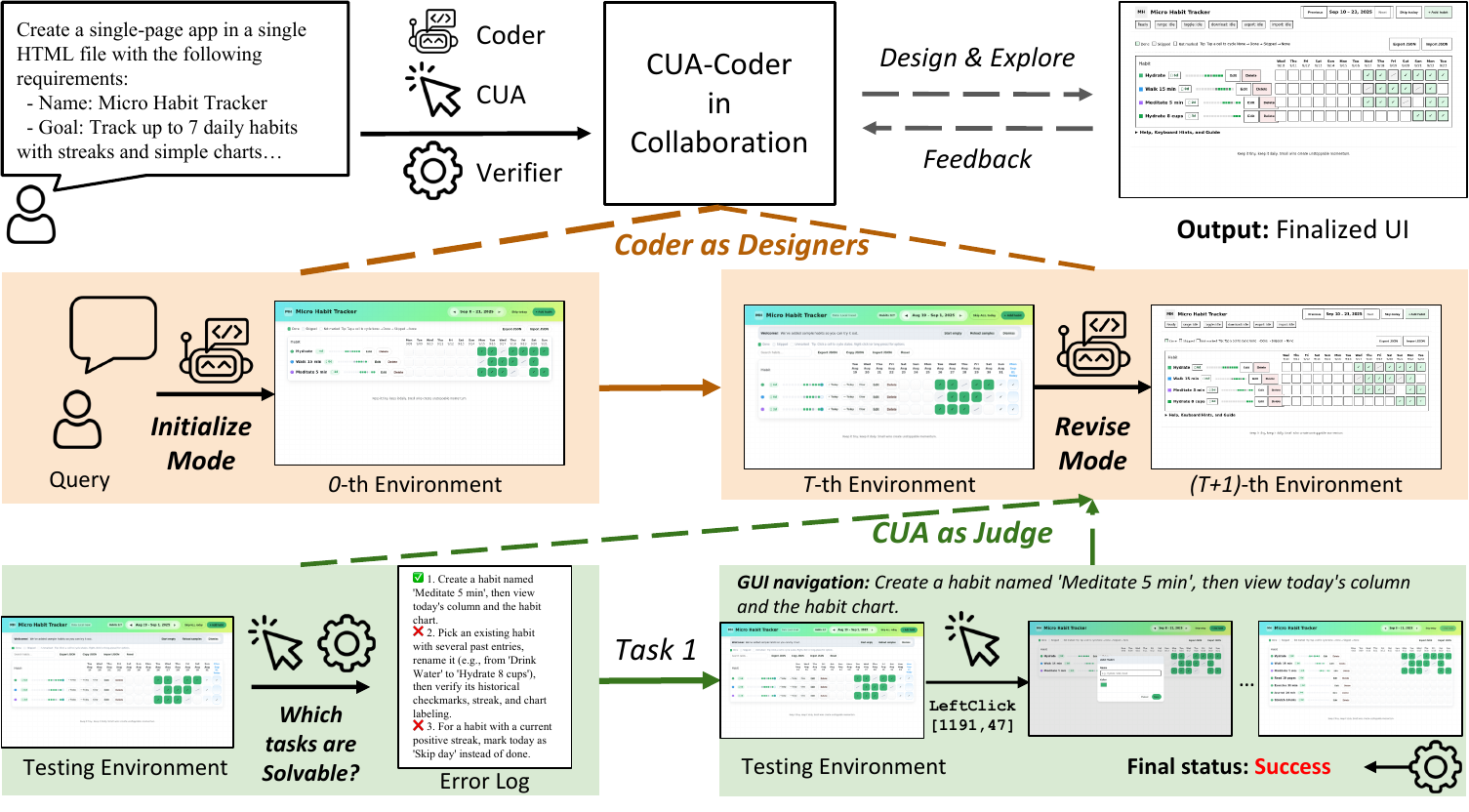}
    \caption{\textbf{Overview of the Coder-CUA in Collaboration framework}. The process begins with the Coder as Designer, which initializes and iteratively revises the UI based on queries and feedback. In parallel, the CUA as Judge executes task-driven navigation within the testing environment, generating trajectories and error logs to evaluate task solvability. A verifier ensures functional correctness, while feedback from CUA navigation informs subsequent UI revisions. This collaboration yields a finalized agent-centric UI optimized for both functionality and execution success.}
    \label{fig:pipeline}
    \vspace{-2em}
\end{figure}

\textbf{Overview.} We present our framework for enabling collaboration between the CUA and the Coder, consisting of two main components: {the Coder as Designer while the CUA as Judge}.
Given a user instruction $\mathcal{Q}$, AUI generates an initial UI environment $\mathcal{E}_0$, which is iteratively revised through interaction and feedback. 
The framework involves two central roles: a {Coder} policy $\pi_{\text{Coder}}$ that proposes and revises UI designs, and a {CUA} policy $\pi_{\text{CUA}}$ that explores the UI and evaluates its functionality. 
We formalize this process as a {Markov Design Process}. 
The {state} is the current UI $\mathcal{E}_t$, the {action} is a design update proposed by $\pi_{\text{Coder}}$, and the {transition} deterministically 
$
\mathcal{E}_{t+1} \leftarrow \pi_{\text{Coder}}(\mathcal{E}_t, \mathcal{R}_t).
$
The {feedback} $\mathcal{R}_t$ is related to the metrics (\ie~Eq.\ref{eq:sr} and Eq.\ref{eq:fc}) results achieved by the CUA when interacting with $\mathcal{E}_t$, \ie~$\mathcal{R}_t \leftarrow S(\mathcal{E}_t,\,\pi_{\text{CUA}}).$ 
The Coder is optimized to maximize the total reward $\mathbb{E}\Big[\sum_t \gamma^t \mathcal{R}_t\Big]$.
In this formulation, the CUA acts as a user that provides actionable feedback by testing the environment, while the Coder serves as a designer who integrates this feedback into code revisions to iteratively improve the UI.
Unlike conventional CUA setups, where the agent adapts to a fixed environment $\pi_{\text{CUA}} \leftarrow \mathcal{E}$, our framework adapts the environment itself based on CUA feedback $\mathcal{E} \leftarrow \pi_{\text{CUA}}$, thereby optimizing UIs for agent-native success. 
We illustrate the full workflow in Fig.~\ref{fig:pipeline} and detail each role in the following subsections.

\vspace{-1em}

\subsection{Coder as Designers}

\vspace{-1em}

Recent advances in Coder~\cite{gpt5blog,qwen3coder,claude4intro} demonstrate strong capabilities in generating UI applications. 
In our framework, we position Coders as \emph{designers}, responsible not only for creating new environments but also for refining them based on feedback from CUAs. 
Accordingly, Coders operate in two complementary modes: one dedicated to the initial creation of UIs, and the other focused on their iterative improvement through CUA-guided feedback.

\textbf{i. Initialization.}  
Given a user query defined in formulation~\ref{prompt} and enriched with multiple details, the Coder progressively generates long-context code to construct a complete HTML-rendered UI $\mathcal{E}_0$ from scratch, which serves as the base environment for subsequent interactions.

\textbf{ii. Revision from Feedback.}  
After constructing the initial environment $\mathcal{E}_0$, the Coder enters an iterative refinement loop to update the UI:  
$\mathcal{E}_{t+1} \leftarrow (\mathcal{E}_t, \mathcal{R}_t),$  
where $\mathcal{R}_t$ denotes the feedback signal expressed as a language caption, described in the next section.

\subsection{CUA as Judges}
We employ Computer-Use Agents (CUAs) as \emph{Judges} to trial and diagnose the UIs $\mathcal{E}_t$ generated by the Coder, providing actionable feedback for iterative redesign. 
Specifically, we define two complementary forms of reward signals:

\textbf{(\textit{i}) Task Solvability Feedback $\mathcal{R}_{\text{task}}$.}
Before navigation begins, we verify whether a task $\tau$ is implementable on the current UI. 
Let $\mathcal{V}$ denote the verifier in Sec.~\ref{sec:verifier}. 
A task is deemed solvable if and only if $\mathcal{V}(\mathcal{E}_t,\tau)=1$; otherwise it is labeled a \emph{functionality failure}. 
This gate prevents wasted rollouts on impossible tasks and sharpens the feedback signal.
We collect all failed tasks into 
$\mathcal{T}_{\text{fail}}=\{\tau:\mathcal{V}(\mathcal{E}_t,\tau)=0\}$ 
and return them to the Coder as precise indicators of missing features.
The Coder then aggregates and summarizes these failures into a language feedback signal $\mathcal{R}_{\text{task}}$.

\begin{table}[!t]
\centering
\footnotesize
\begin{tabular}{m{0.12\linewidth} m{0.82\linewidth}}
\toprule
\centering \textbf{Task} & 
\centering Load the app for the first time and wait for the curtain reveal to complete. \tabularnewline
\midrule
\centering \textbf{Website} & 
\centering \includegraphics[width=0.5\linewidth]{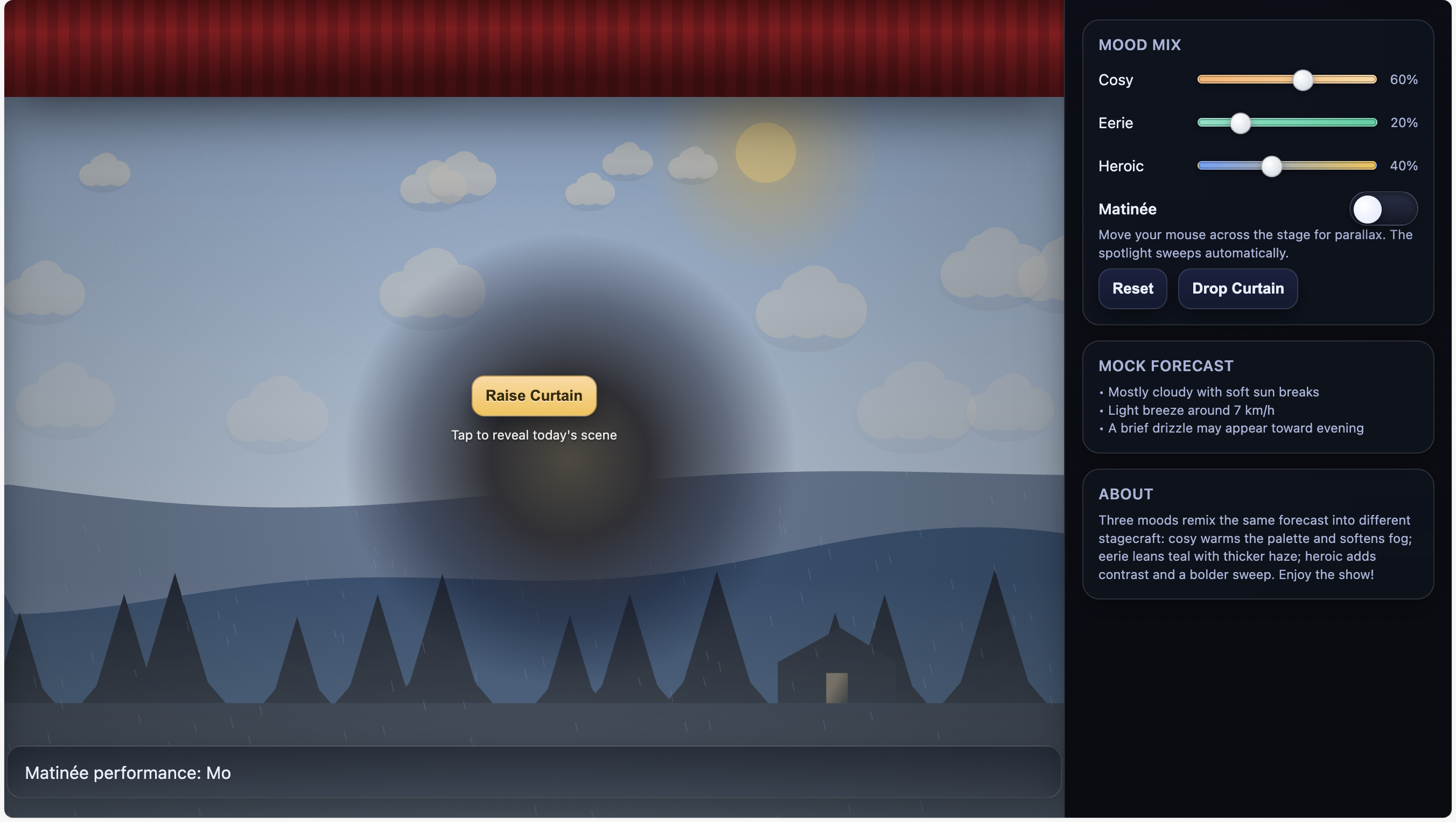}\\
$1280\times 720$\tabularnewline
\midrule
\centering \textbf{Dashboard}\\(an image) & 
\centering \includegraphics[width=\linewidth]{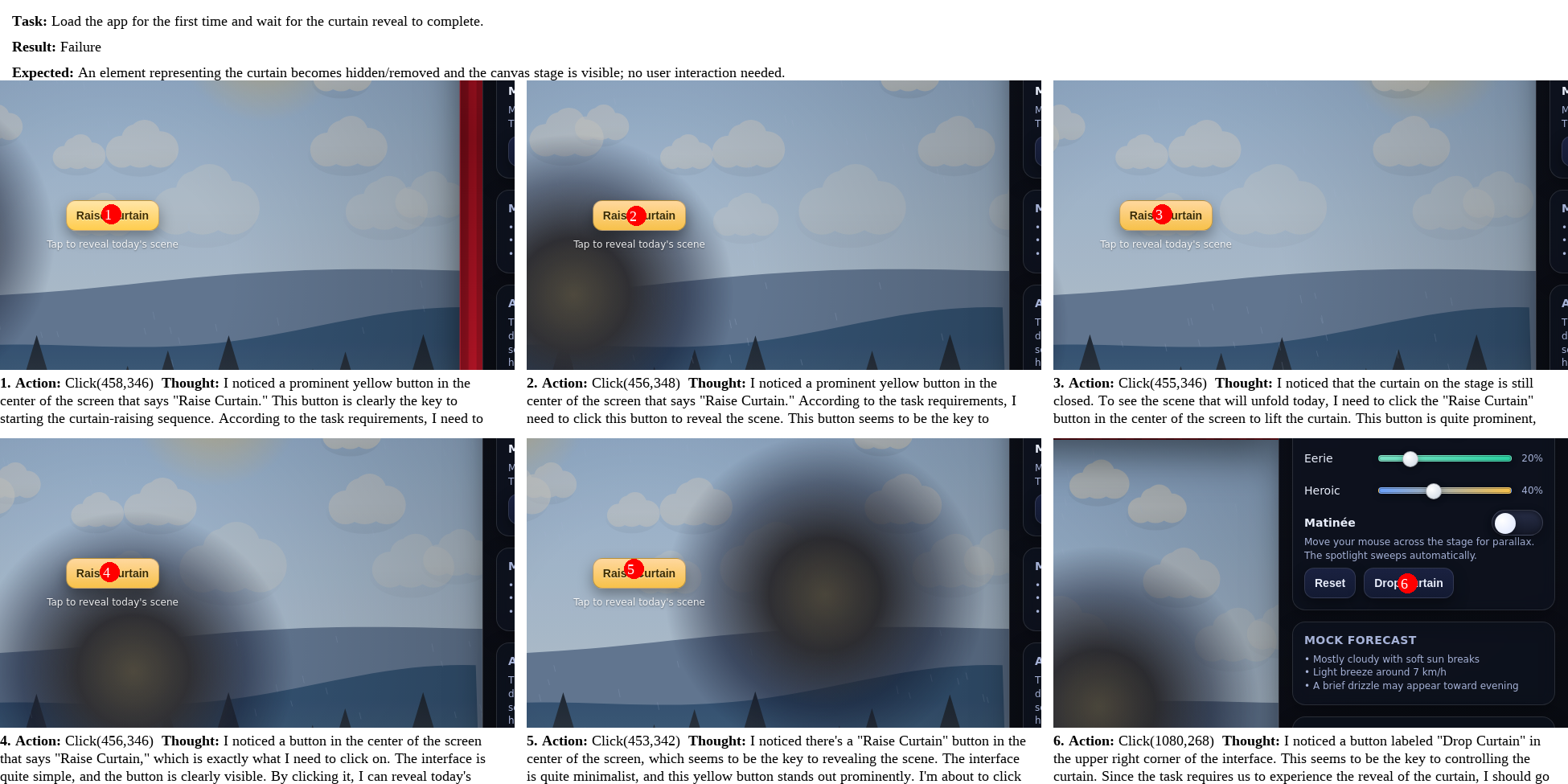}\\
Before: $6\times 1280\times 720$ $\rightarrow$ After: $1\times 1950\times 975$, 76.2\% tokens reduction \tabularnewline
\midrule
\centering \textbf{Result} & 
\centering Failure \tabularnewline
\midrule
\centering \textbf{Comments} & 
\centering The weather-theatre app requires  button clicks to trigger curtain reveal, 
but the task expects automatic curtain opening on first load without user interaction, 
{creating a fundamental mismatch between expected auto-start behavior and actual manual activation requirement}. 
\tabularnewline
\bottomrule
\end{tabular}
\caption{\textbf{Illustration of CUA Dashboard.} The dashboard generates one informative image that clearly demonstrates how the CUA performs each step along with the corresponding observations, while reducing visual tokens by cropping to the key interactive regions.}
\label{tab:dashboard}
\vspace{-1em}
\end{table}

\textbf{(\textit{ii}) CUA Navigation Feedback $\mathcal{R}_\text{nav}$.}
For solvable tasks $\mathcal{T}_{\text{succ}}=\{\tau:\mathcal{V}(\mathcal{E}_t,\tau)=1\}$ , evaluation proceeds as a UI navigation problem. 
At step $k$, the CUA receives an observation $o_k$ (a screenshot of the current state), emits an action $a_k\in\{\textsc{click},\textsc{type},\textsc{scroll},\ldots\}$ with an optional reasoning trace, and the environment transitions to the next state, yielding $o_{k+1}$. 
The trajectory $\mathcal{H}=\left(o_0,a_0,\ldots,o_K\right)$ terminates when either (a) the function checker signals success $\widetilde{\mathcal{V}}(\mathcal{E}_t,\tau)=1$, or (b) a step limit is reached, which we record as a {failure}. 
We log full trajectories---observations, actions, and intermediate rationales---and use them to construct targeted feedback for UI refinement.

\textbf{CUA Dashboard for Compact Feedback.}
Raw trajectories $\mathcal{H}$ are long and interleaved, making them ill-suited for direct ingestion by the Coder. 
We therefore distill each rollout into a {CUA Dashboard} (Fig.~\ref{tab:dashboard}): a single, fixed-resolution ($1920\times1080$) canvas that compresses key evidence from the trial. 
Rather than storing full frames, we crop and tile only \emph{interactive regions} touched by the CUA, allocating dynamic region sizes based on step order to preserve temporal structure. 
This yields a substantial reduction in redundancy (\eg~a $76.2\%$ drop in visual content) while retaining the cues needed to localize failure modes (missed affordances, hidden state, ambiguous labels) and success paths at a glance. 
The dashboard provides a step-by-step visual trace aligned with actions, making error locations immediately visible.
Finally, we convert the dashboard into a concise language summary $\mathcal{R}_\text{nav}$ by passing it to a VLM as Commenter then as the feedback for revision.

%% file: section/exps.tex
\vspace{-1em}

\section{Experiments}
\subsection{Settings}
For the Coder, we evaluate GPT-5~\cite{gpt5blog}, GPT-4o~\cite{gpt4o}, and the open-source Qwen3-Coder~\cite{qwen3coder}. For the CUA, we use UI-TARS-1.5-7B~\cite{uitars}, a lightweight yet efficient open model, and Operator~\cite{openaioperator}, a state-of-the-art closed-source API-based CUA. GPT-5 serves both as Coder and Commenter, configured with high verbosity and reasoning effort for coding, and low verbosity with minimal reasoning for commenting. GPT-4o also serves as both Coder and Commenter but without specific verbosity or reasoning configurations. For Qwen, coding is performed by Qwen3-Coder-30B-A3B-Instruct, while commenting is done by Qwen2.5-VL-72B-Instruct. The Task Proposer and Verifier utilize GPT-5 with high verbosity and reasoning effort. In CUA policy tests, we limit maximum steps to 20 to prevent infinite loops, conducting evaluations using Playwright. CUAs exclusively perform coordinate-based Computer Use actions without direct interaction with UI elements, enhancing evaluation difficulty and providing deeper insights into UI layout and visibility. Experiment results consistently demonstrate universal performance gains from our proposed method, benefiting CUAs of varying complexity.

\begin{table*}[!t]
\centering
\caption{\textbf{Main results on \bench~per Coder.} Top: {Func. Completeness Rate} (\%). Bottom: {CUA Success Rate} (\%).}
\label{tab:unified_uitars_three_models_split}
\resizebox{\linewidth}{!}{%
\begin{tabular}{llccccccc}
\toprule
\textbf{Coder} & \textbf{Feedback Type} & \textbf{landing (\%)} & \textbf{game (\%)} & \textbf{app (\%)} & \textbf{utility (\%)} & \textbf{interactive (\%)} & \textbf{tool (\%)} & \textbf{overall (\%)} \\
\midrule
\multicolumn{9}{c}{\textcolor{gray}{\textit{Function Completeness}}} \\
\midrule
\multirow{4}{*}{GPT-5}
  & Baseline       & 53.0 & 77.8 & 70.6 & 63.3 & 73.0 & 70.0 & 67.9 \\
  & + Task Solvability    & 19.7 & 100.0& 69.4 & 65.6 & 55.6 & 56.2 & 60.5 \\
  & + CUA Navigation   & 53.3 & 87.8 & 74.2 & 70.0 & 70.4 & 69.5 & 70.8 \\
  & + Integrated   & 75.3 & 92.2 & 85.2 & 73.3 & 82.6 & 76.7 & \textbf{81.5} \\
\midrule
\multirow{4}{*}{\makecell[l]{Qwen3-\\Coder-30B}}
  & Baseline       & 16.3 & 50.4 & 41.2 & 43.9 & 52.2 & 54.8 & 42.1 \\
  & + Task Solvability   & 55.0 & 79.6 & 58.5 & 67.8 & 56.3 & 74.3 & \textbf{64.3} \\
  & + CUA Navigation   & 23.3 & 50.4 & 38.8 & 49.4 & 39.3 & 55.2 & 41.3 \\
  & + Integrated   & 47.7 & 72.2 & 59.7 & 56.7 & 57.0 & 69.5 & 60.1 \\
\midrule
\multirow{4}{*}{GPT-4o}
  & Baseline       & 9.7  & 55.2 & 36.1 & 38.9 & 44.8 & 37.6 & 36.3 \\
  & + Task Solvability    & 23.7 & 55.9 & 52.1 & 55.0 & 58.9 & 65.2 & \textbf{50.6} \\
  & + CUA Navigation    & 8.3  & 55.2 & 28.2 & 34.4 & 26.3 & 35.7 & 30.4 \\
  & + Integrated   & 16.3 & 68.5 & 36.4 & 51.7 & 51.1 & 41.4 & 43.1 \\
\midrule
\multicolumn{9}{c}{\textcolor{gray}{\textit{CUA Success Rate}}} \\
\midrule
\multirow{4}{*}{GPT-5}
  & Baseline       & 34.7 & 24.8 & 27.3 & 14.4 & 18.1 & 21.9 & 24.5 \\
  & + Task Solvability    & 16.3 & 39.3 & 26.7 & 16.1 & 20.7 & 11.9 & 22.6 \\
  & + CUA Navigation    & 17.7 & 43.3 & 30.0 & 21.1 & 21.1 & 17.6 & 25.7 \\
  & + Integrated   & 40.7 & 27.4 & 31.5 & 22.2 & 14.1 & 12.9 & \textbf{26.0} \\
\midrule
\multirow{4}{*}{\makecell[l]{Qwen3-\\Coder-30B}}
  & Baseline       & 5.3  & 9.3  & 9.1  & 11.7 & 7.0  & 1.4  & 7.3  \\
  & + Task Solvability    & 14.7 & 42.2 & 19.1 & 14.4 & 11.1 & 4.3  & 18.3 \\
  & + CUA Navigation    & 6.7  & 20.7 & 9.1  & 11.1 & 12.2 & 11.4 & 11.7 \\
  & + Integrated   & 23.7 & 30.7 & 22.4 & 7.8  & 9.3  & 13.8 & \textbf{19.0} \\
\midrule
\multirow{4}{*}{GPT-4o}
  & Baseline       & 4.7  & 12.6 & 12.4 & 6.7  & 9.3  & 5.7  & 8.8  \\
  & + Task Solvability    & 8.7  & 18.5 & 19.1 & 5.6  & 8.5  & 22.9 & 14.1 \\
  & + CUA Navigation   & 5.7  & 31.5 & 10.0 & 8.3  & 10.4 & 6.7  & 12.3 \\
  & + Integrated   & 10.3 & 27.4 & 13.9 & 13.3 & 15.2 & 16.7 & \textbf{16.1} \\
\bottomrule
\end{tabular}%
}
\vspace{-2em}
\end{table*}

\subsection{Main Results}
Table~\ref{tab:unified_uitars_three_models_split} reports results across six domains for three coders. Several key findings emerge:
\textbf{(i) Function Completeness.} Revision based on task solvability feedback leads to substantial gains, consistently boosting the overall functionality completeness for all coders. After applying integrated revision for GPT-5, the function completeness is increased to 81.5\% from 67.9\%, reaching the highest. Notably, the landing, game and app domains have dramatic improvements, with the maximum improvements of 31.4\%. Interestingly, revision based on task solvability feedback or CUA navigation feedback alone does not guarantee function completeness improvements, but the integrated revision combining these two components bring stable improvements in all domains for all coders, highlighting the strength of our design. Moreover, fixing unresolved functionalities alone also benefits CUA task solving, yielding a 4.8\% average improvement on CUA evaluation, highlighting the mutual reinforcement between task solvability and CUA navigation.

\textbf{(ii) CUA performance.} Open-source CUAs initially perform poorly, with an average overall CUA success rate of only 13.5\%. However, our framework can consistently improve the CUA success rate, with an average 6.8\% improvements. Interestingly, our framework brings large improvements to weak coders such as Qwen3-Coder-30B and GPT-4o, with a maximum overall improvement of 11.7\%, showcasing that our framework can greatly empower weak models.
Overall, these results demonstrate the effectiveness of our framework: task solvability feedback guides to robust UI design, while leveraging CUA navigation feedback optimizes interfaces toward agent-centric success.

\subsection{Ablation Studies}

\begin{figure*}[!h]
  \centering
  \vspace{-1em}
  \setlength{\tabcolsep}{0pt}
  \renewcommand{\arraystretch}{1.0}
  \begin{tabular}{cccc}
    \subcaptionbox{Func. completeness.\label{fig:commenter-func}}{
      \includegraphics[width=0.24\textwidth]{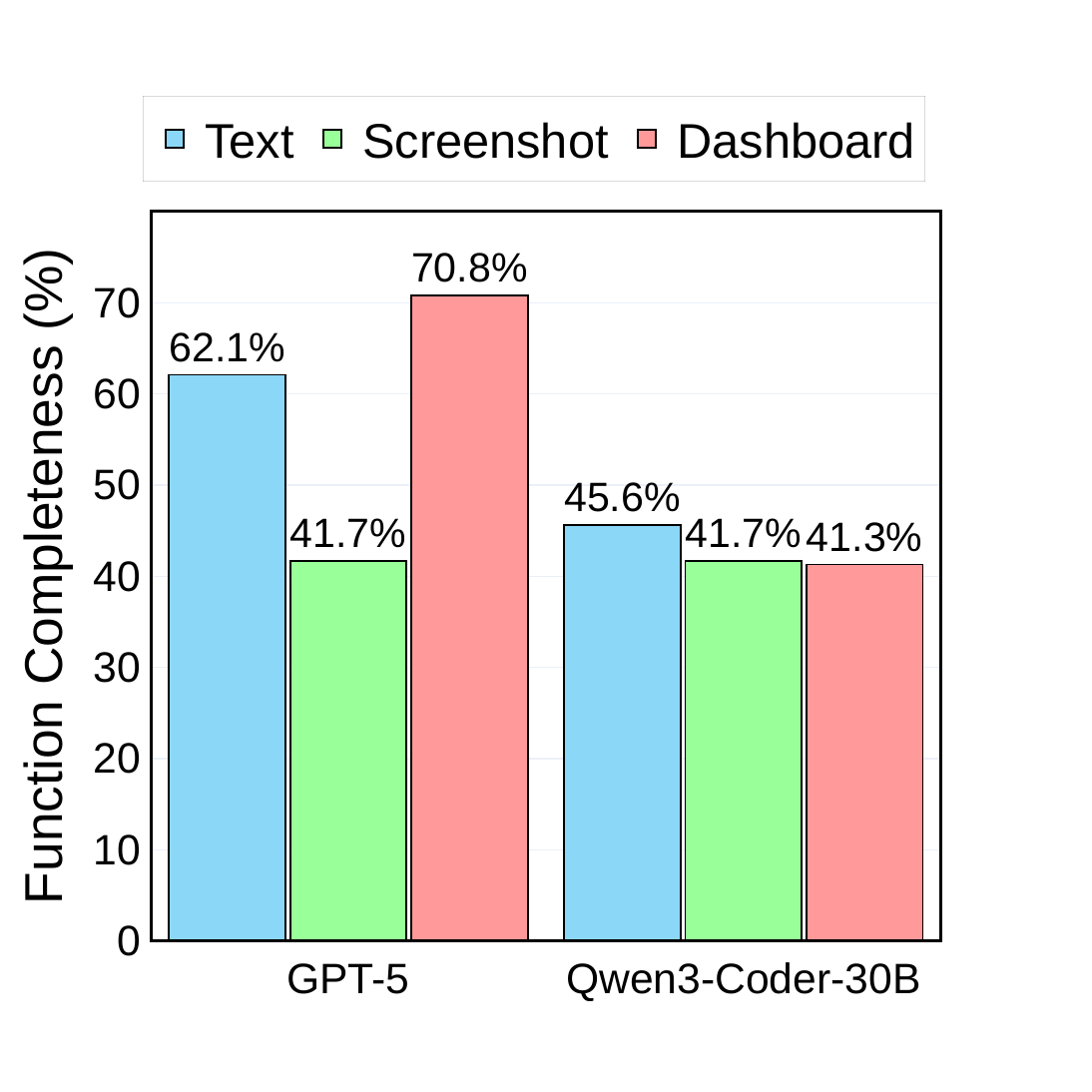}} &
    \subcaptionbox{CUA success rate.\label{fig:commenter-cua}}{
      \includegraphics[width=0.24\textwidth]{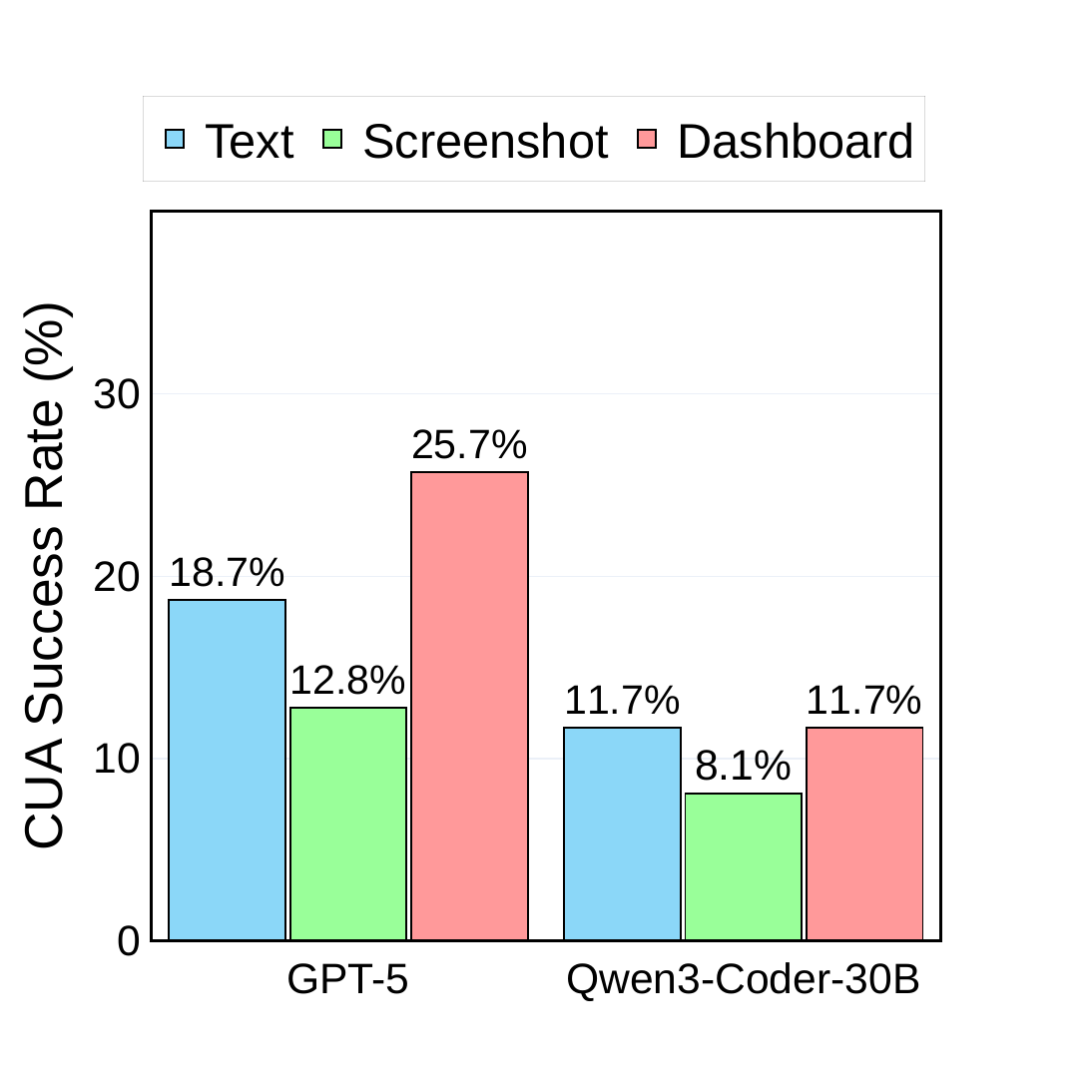}} &
    \subcaptionbox{Func. completeness.\label{fig:iterative-func}}{
      \includegraphics[width=0.24\textwidth]{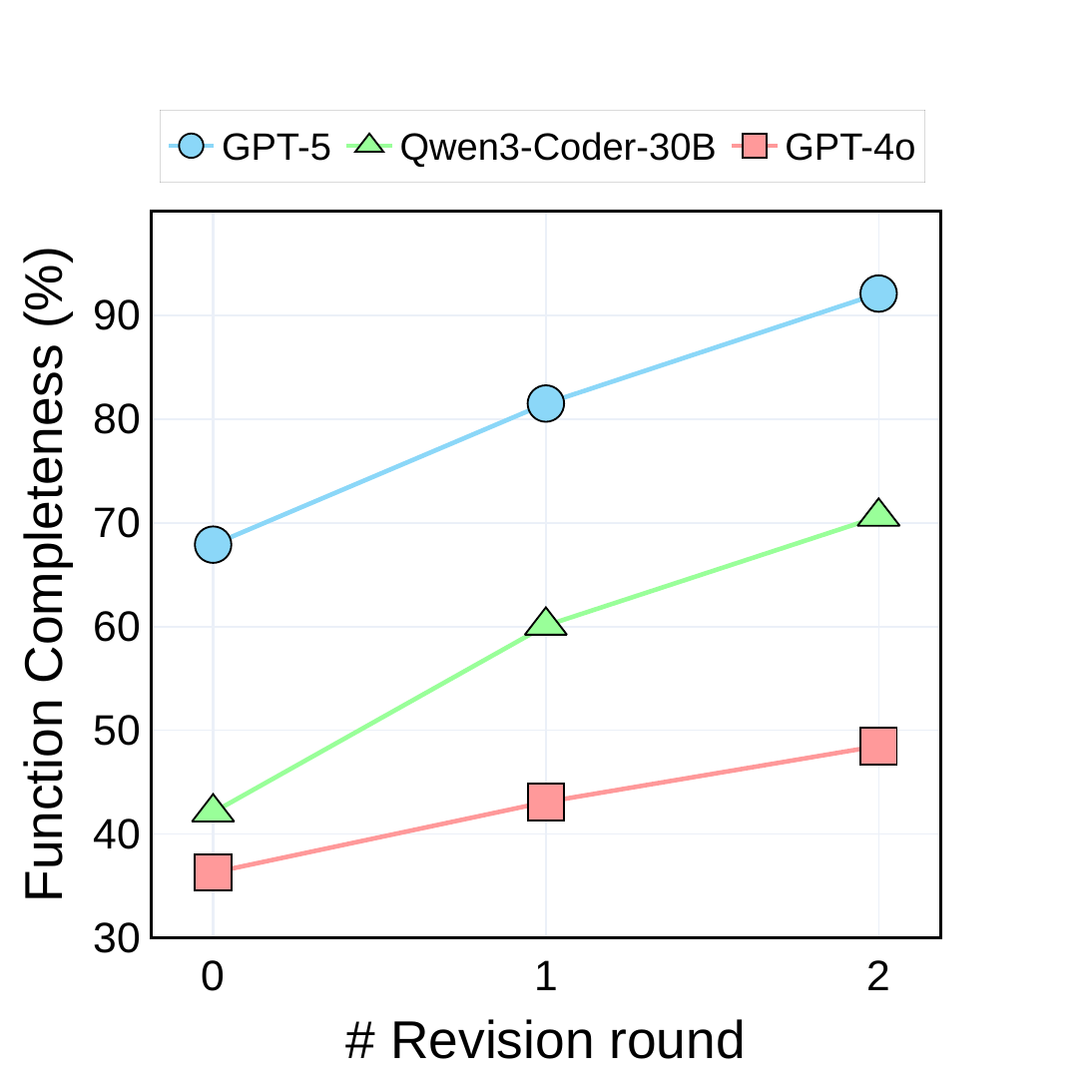}} &
    \subcaptionbox{CUA success rate.\label{fig:iterative-cua}}{
      \includegraphics[width=0.24\textwidth]{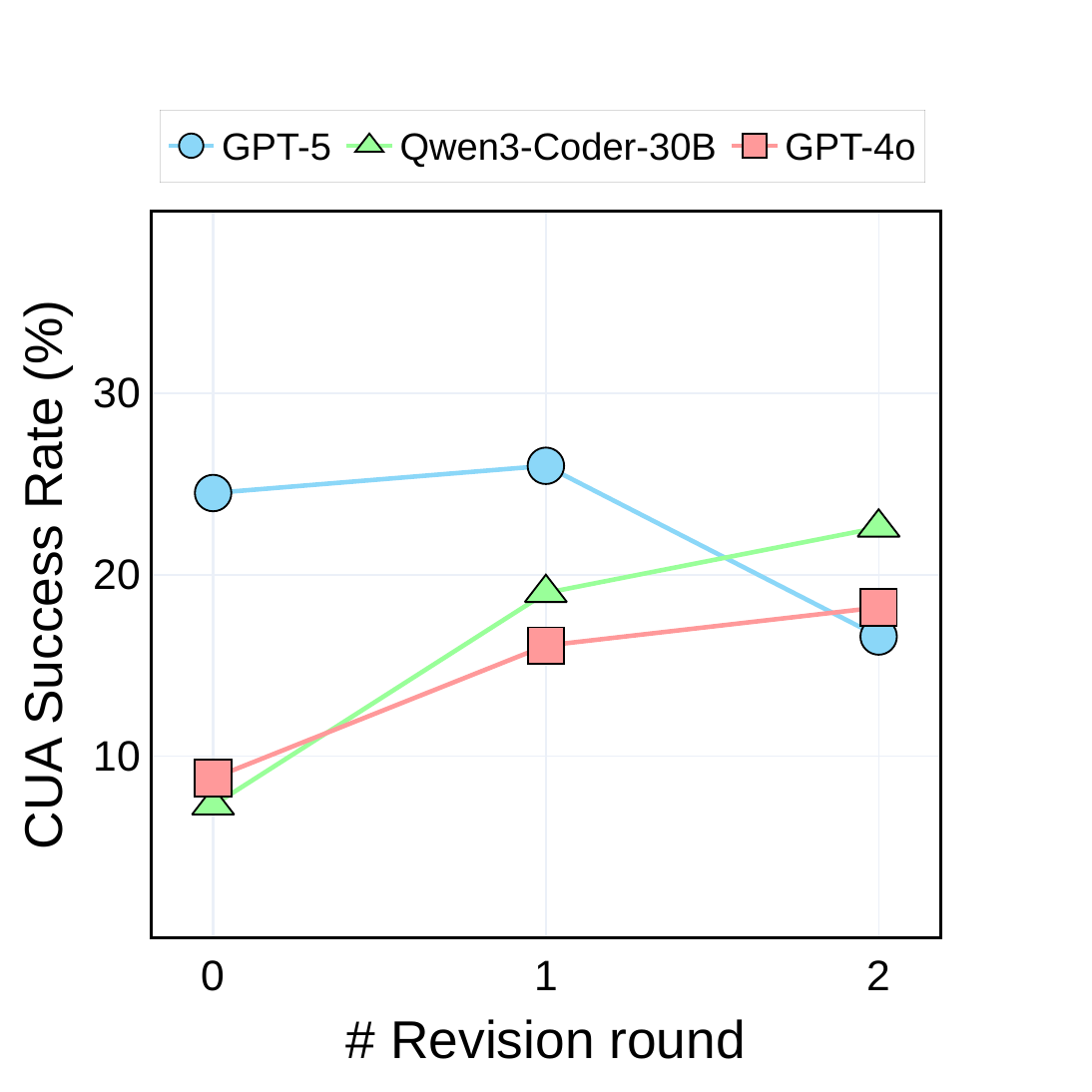}}
  \end{tabular}

  \caption{\textbf{Ablation Studies of CUA Dashboard and Iterative rounds.}
  Left (a-b): Effects by CUA Dashboard.
  Right (c-d): Performance across different iterative revision rounds.}
  \label{fig:abl:merged}
\end{figure*}

\textbf{Ablation studies of CUA Dashboard.} 
To ablate the Dashboard, we design two commenter variants: one using textual actions only, and the other using visual screenshots only. We evaluate both variants (Coder, Commenter) under the setting that (GPT-5, GPT-5) and (Qwen3-Coder-30B, Qwen2.5-VL-72B). As shown in Fig.~\ref{fig:abl:merged}(a-b), our Dashboard consistently brings gains, significantly improving both Function Completeness (from 62.1\% to 70.8\%) and CUA Success Rate (from 18.7\% to 25.7\%) compared to the action variant for GPT-5. In contrast, the screenshot variant mostly performs worst, highlighting that visual inputs solely are inadequate for providing refinement insights, while integrating visual and textual information in our Dashboard notably benefits the commenting process.

\textbf{Effects by Refinement Round.} As shown in the Fig.\ref{fig:abl:merged}(c-d), iteratively applying revision can consistently bring gains on the function completeness for all coders. Interestingly, it can be observed that the CUA success rate of GPT-5 coder may drop after repeated revision, while Qwen3-Coder-30B and GPT-4o can consistently gain from repeated revision. This indicates that the revision improvement may saturate for strong coders, but weak coders can be improved with iterative revisions.

\subsection{Qualitative Analysis}

\begin{figure}[!h]
    \centering

    \begin{subfigure}{\textwidth}
    \centering
    \begin{minipage}[b]{0.32\textwidth}
        \centering
        \includegraphics[width=\linewidth]{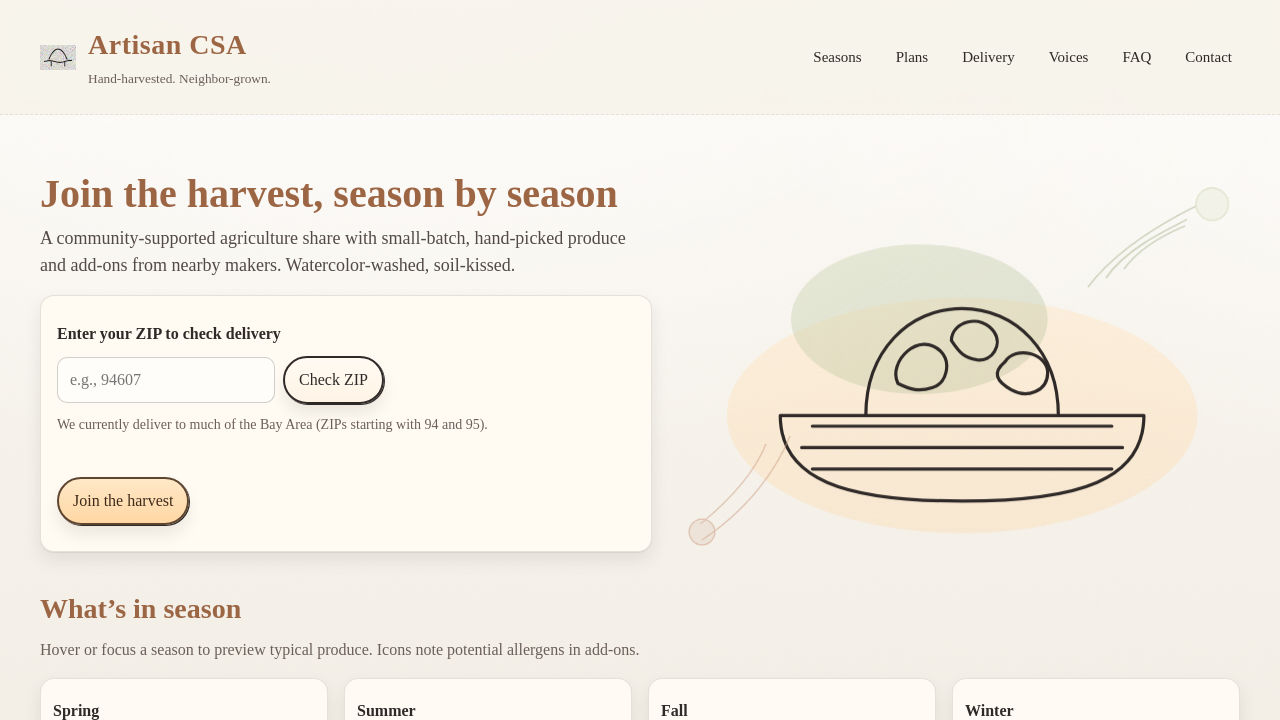}
        \caption*{Initial UI}
    \end{minipage}
    \begin{minipage}[b]{0.32\textwidth}
        \centering
        \includegraphics[width=\linewidth]{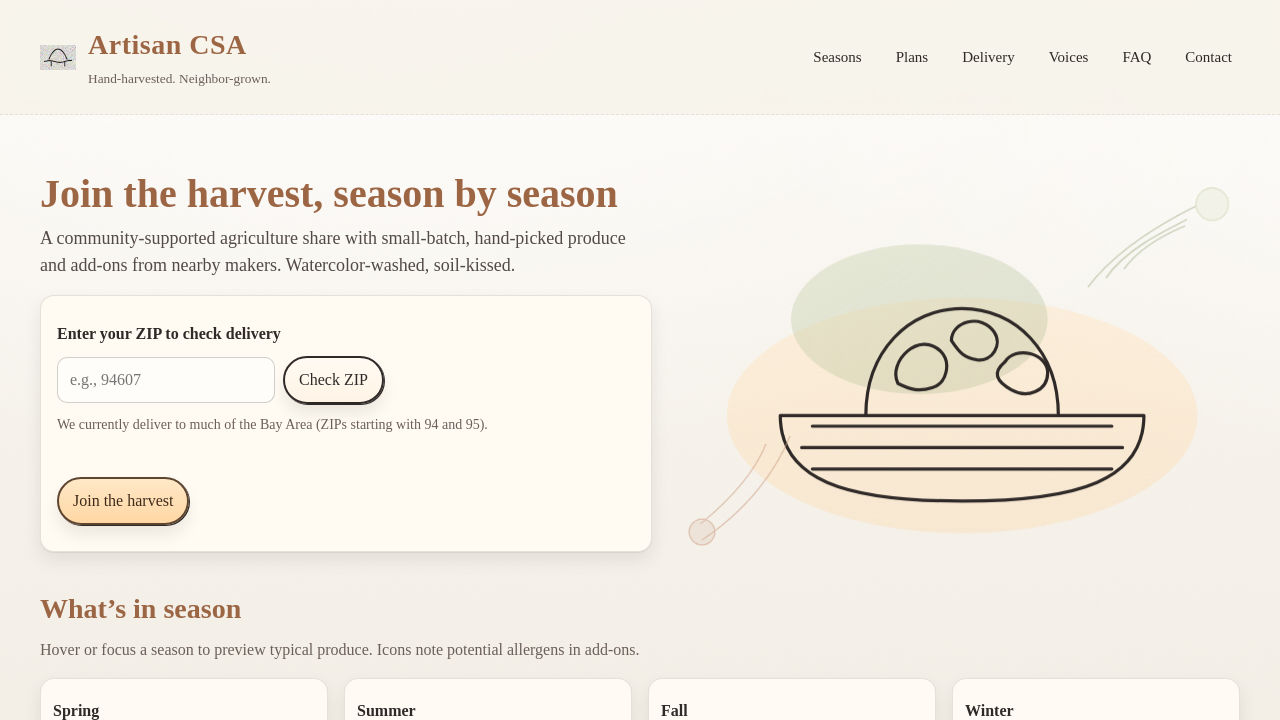}
        \caption*{w. Task Solvability Feedback}
    \end{minipage}
    \begin{minipage}[b]{0.32\textwidth}
        \centering
        \includegraphics[width=\linewidth]{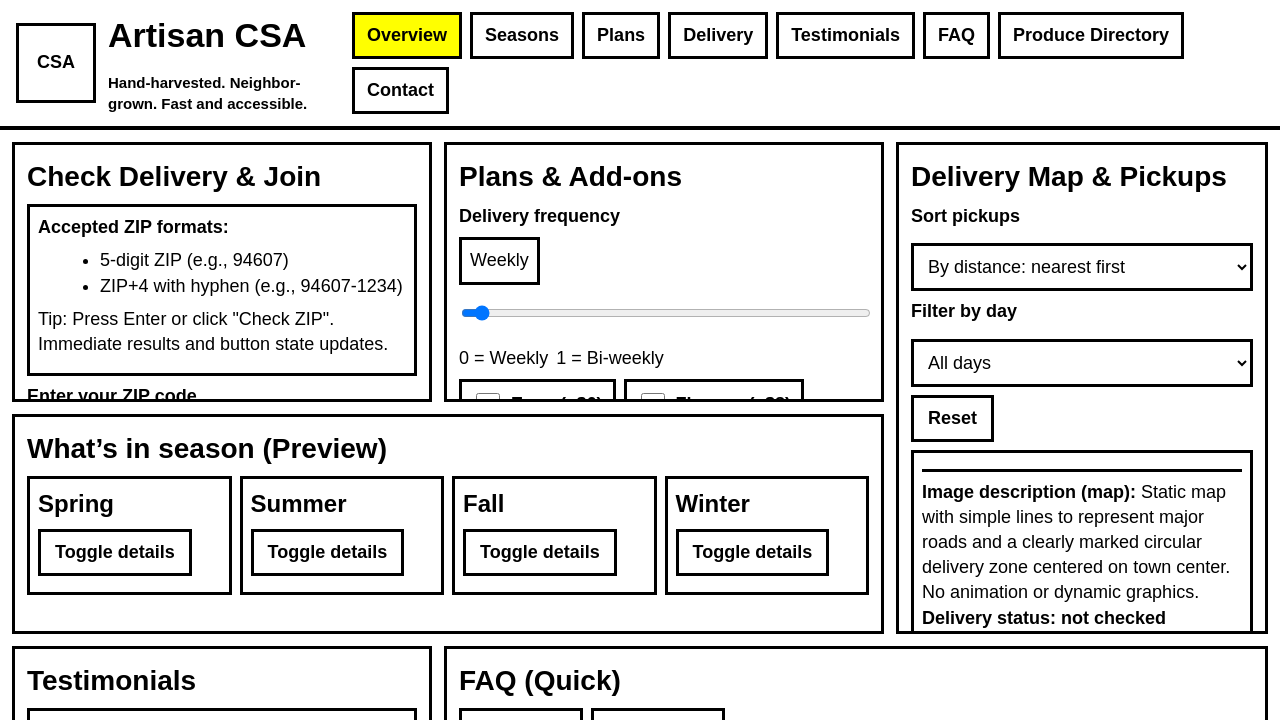}
        \caption*{w. CUA Navigation Feedback}
    \end{minipage}
    \caption{\texttt{artisan-csa}: Create a single-page app, in a single HTML file, for a community-supported agriculture program with a hand-drawn, watercolor aesthetic.}
    \label{fig:qualitative_artisan-csa}
    \end{subfigure}

    \begin{subfigure}{\textwidth}
    \centering
    \begin{minipage}[b]{0.32\textwidth}
        \centering
        \includegraphics[width=\linewidth]{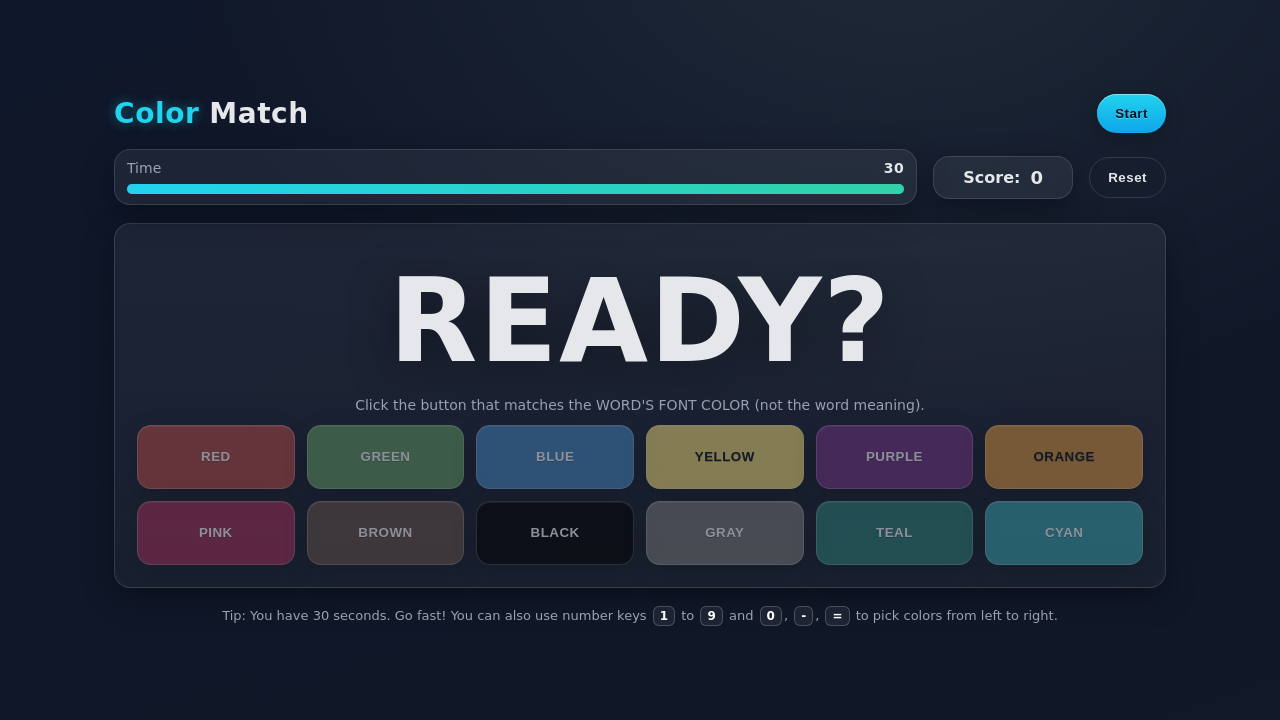}
        \caption*{Initial UI}
    \end{minipage}
    \begin{minipage}[b]{0.32\textwidth}
        \centering
        \includegraphics[width=\linewidth]{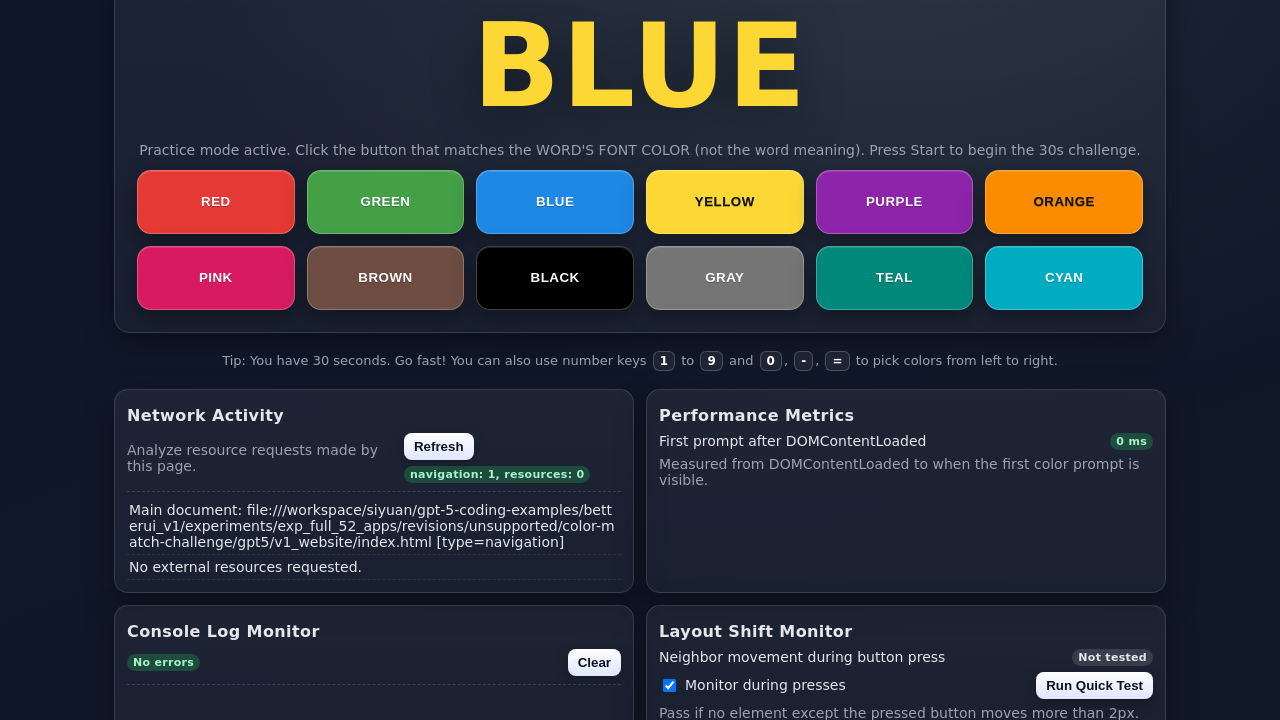}
        \caption*{w. Task Solvability Feedback}
    \end{minipage}
    \begin{minipage}[b]{0.32\textwidth}
        \centering
        \includegraphics[width=\linewidth]{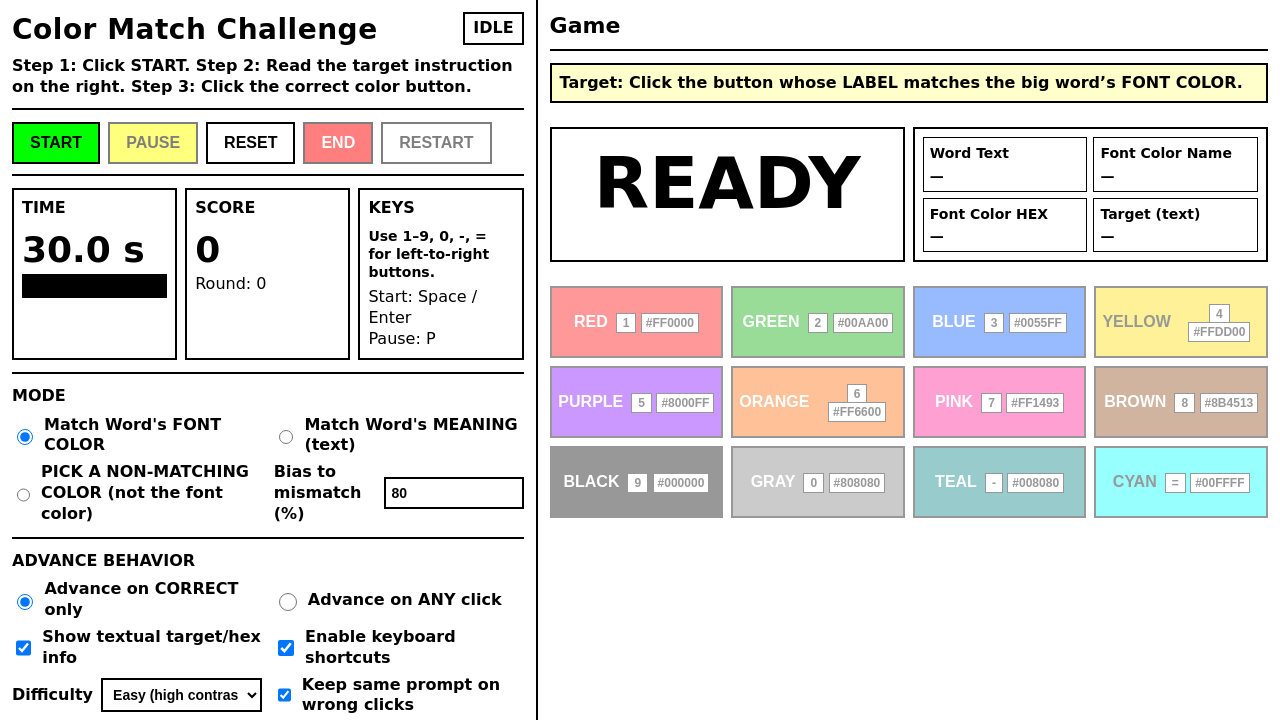}
        \caption*{w. CUA Navigation Feedback}
    \end{minipage}
    \caption{\texttt{color-match-challenge}: Create a single-page app in a single HTML file for a fast-paced “color match” game. - Show a word (e.g., “RED”) in a random font color — player must click the correct color button (not the word meaning). - Keep score based on correct answers within 30 seconds. - Use large typography, color-coded buttons, and smooth button press animations.}
    \label{fig:qualitative_color-match-challenge}
    \end{subfigure}

    \begin{subfigure}{\textwidth}
    \centering
    \begin{minipage}[b]{0.32\textwidth}
        \centering
        \includegraphics[width=\linewidth]{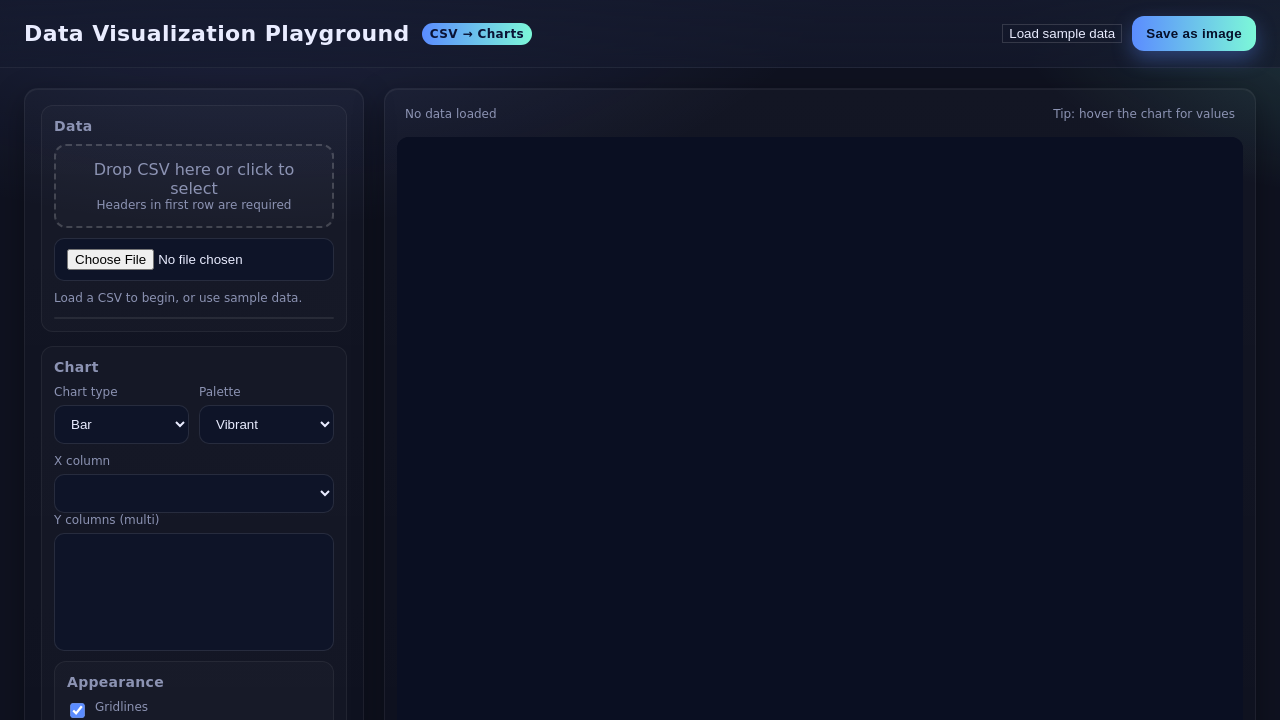}
        \caption*{Initial UI}
    \end{minipage}
    \begin{minipage}[b]{0.32\textwidth}
        \centering
        \includegraphics[width=\linewidth]{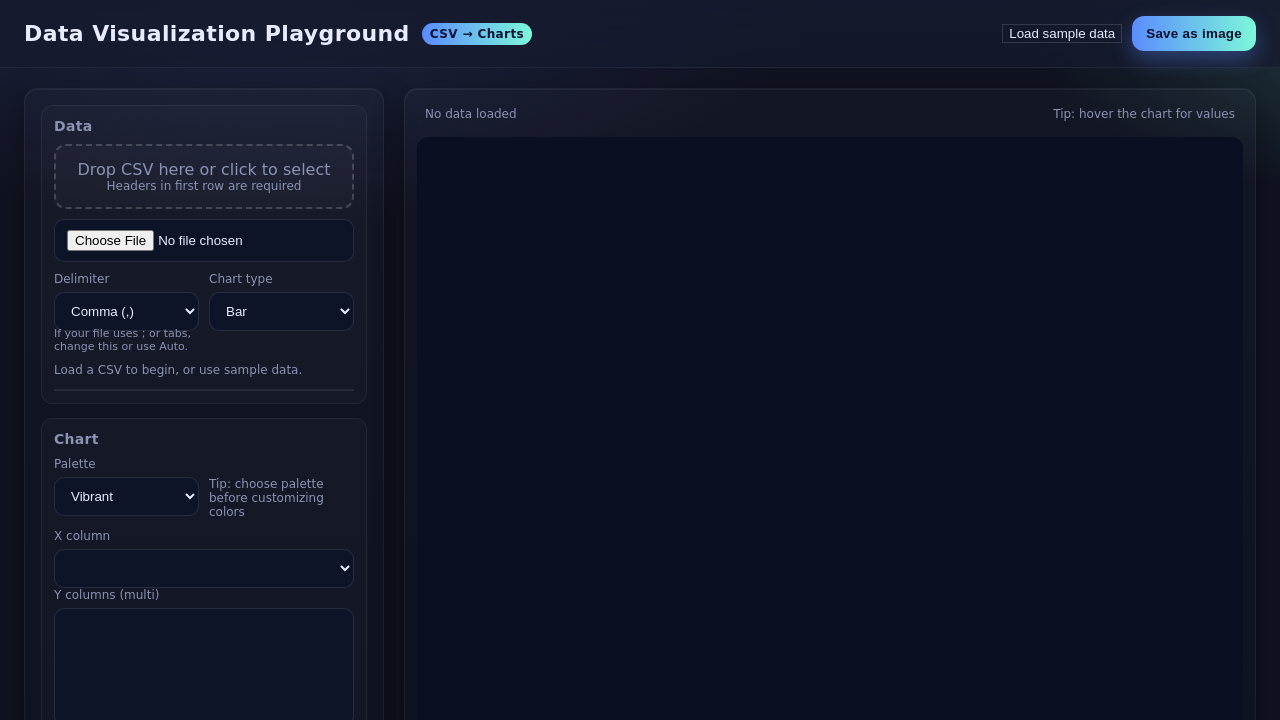}
        \caption*{w. Task Solvability Feedback}
    \end{minipage}
    \begin{minipage}[b]{0.32\textwidth}
        \centering
        \includegraphics[width=\linewidth]{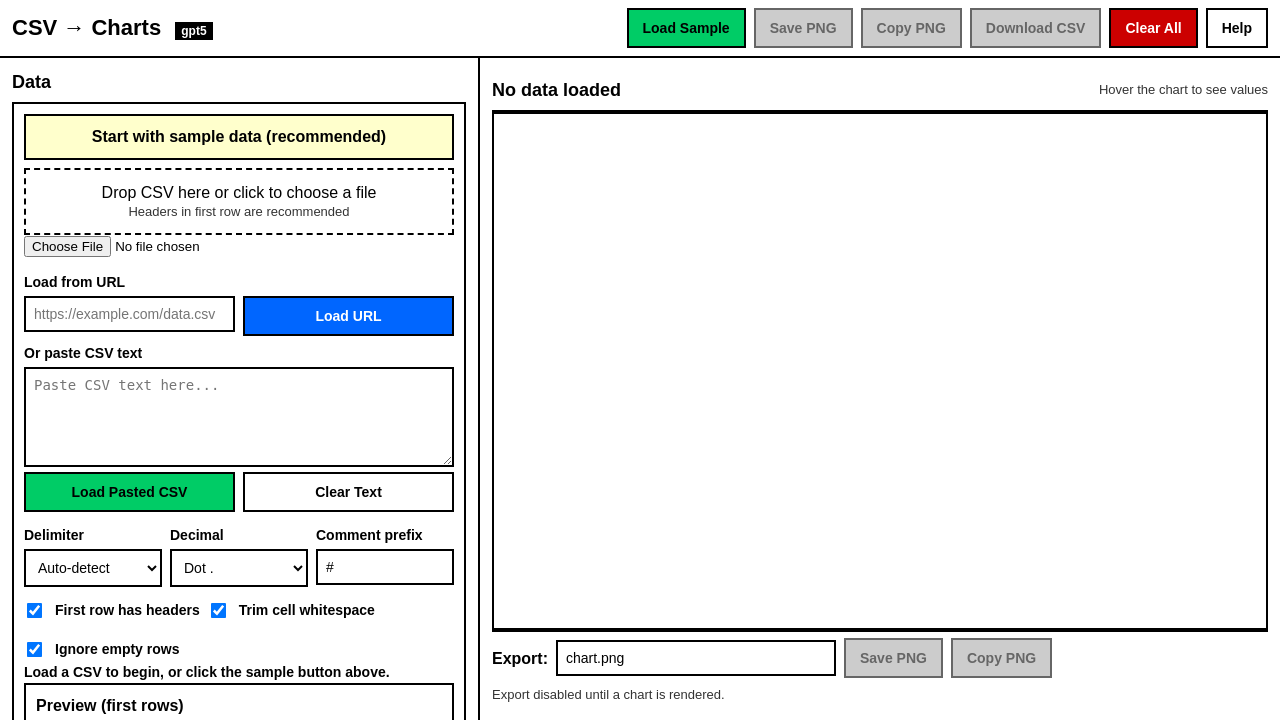}
        \caption*{w. CUA Navigation Feedback}
    \end{minipage}
    \caption{\texttt{csv-to-charts}: Create a single-page app in a single HTML file with the following requirements: - Name: Data Visualization Playground - Goal: Upload CSV and generate charts. - Features: Chart type selector, color customization, save as image. - The UI should be modern with a focus on charts.}
    \label{fig:qualitative_csv-to-charts}
    \end{subfigure}

    \begin{subfigure}{\textwidth}
    \centering
    \begin{minipage}[b]{0.32\textwidth}
        \centering
        \includegraphics[width=\linewidth]{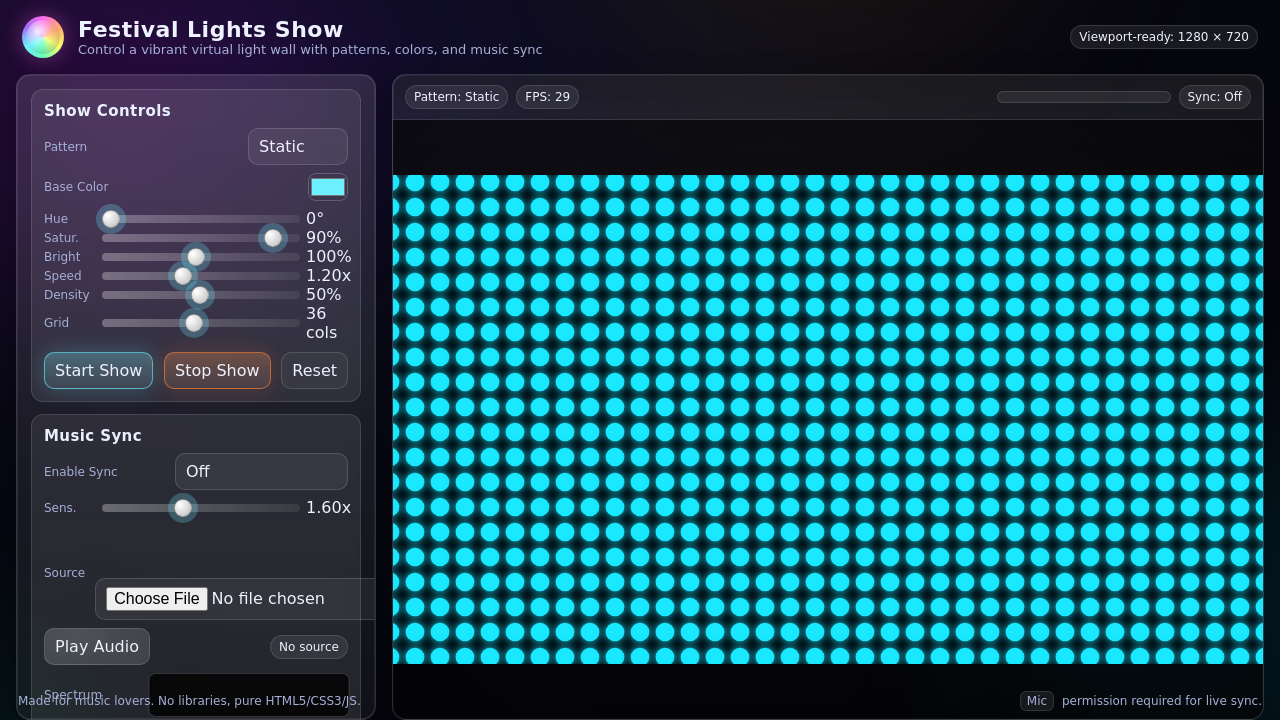}
        \caption*{Initial UI}
    \end{minipage}
    \begin{minipage}[b]{0.32\textwidth}
        \centering
        \includegraphics[width=\linewidth]{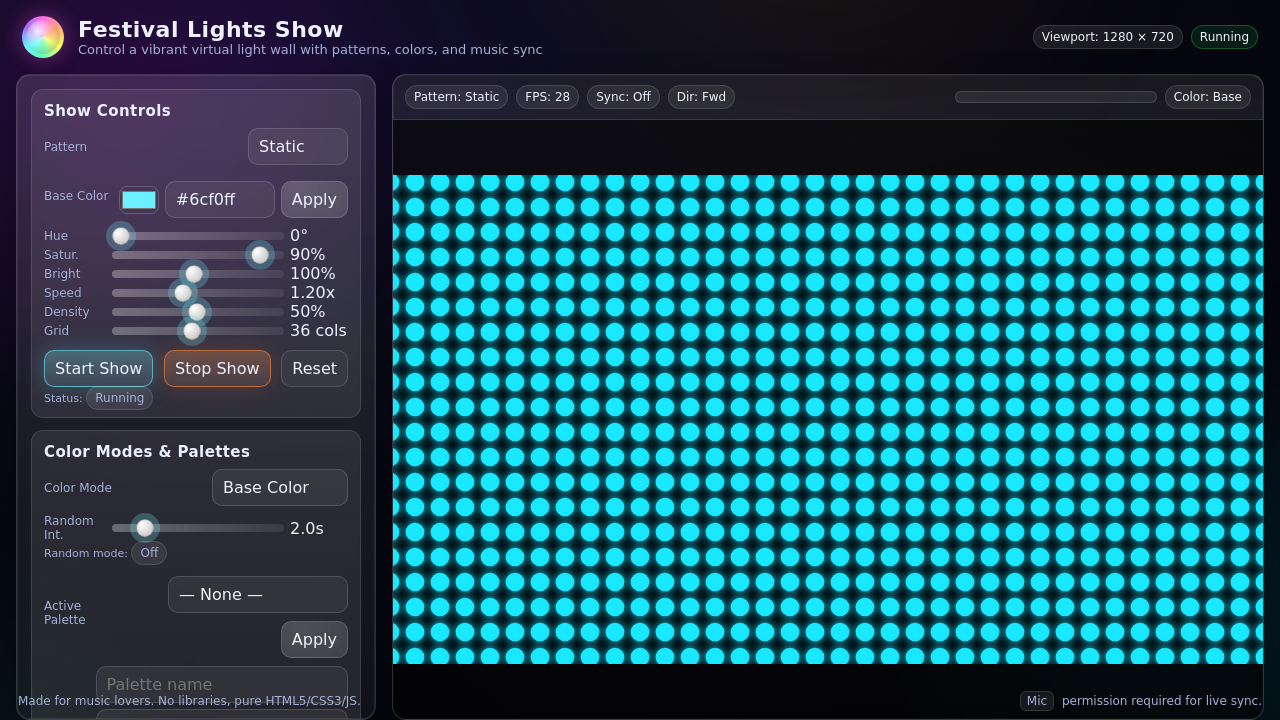}
        \caption*{w. Task Solvability Feedback}
    \end{minipage}
    \begin{minipage}[b]{0.32\textwidth}
        \centering
        \includegraphics[width=\linewidth]{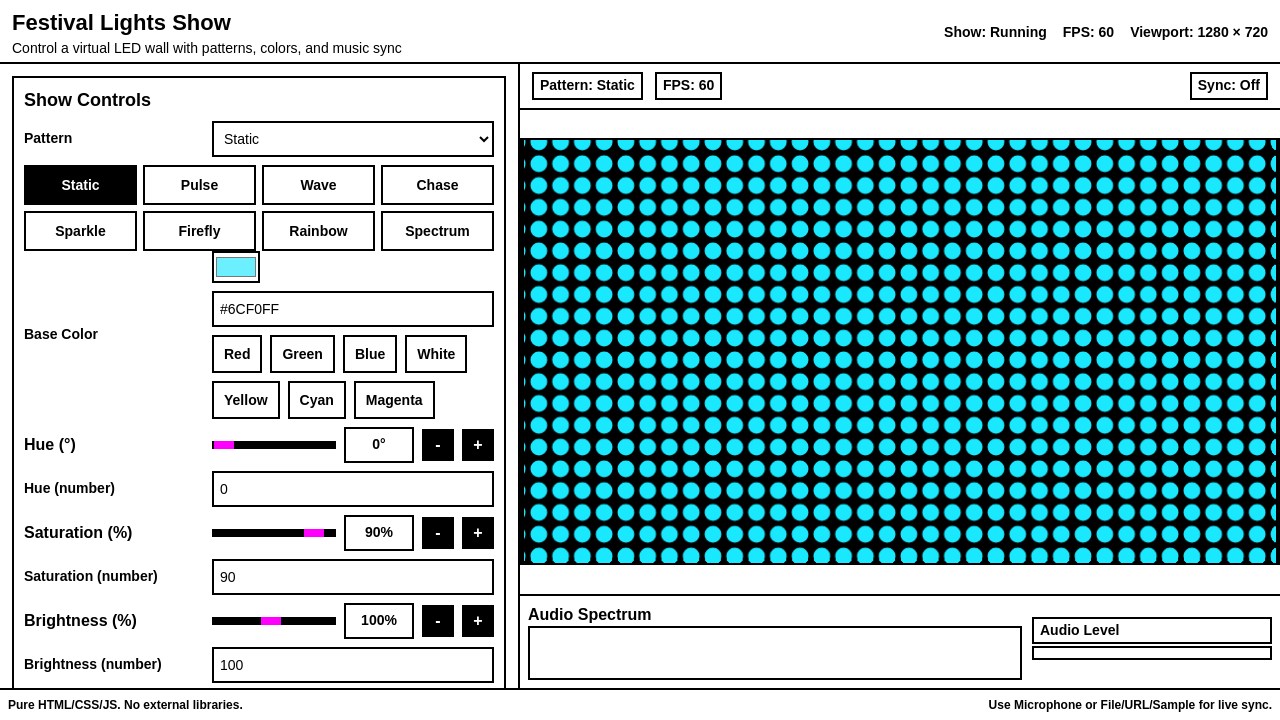}
        \caption*{w. CUA Navigation Feedback}
    \end{minipage}
    \caption{\texttt{festival-lights-show}: Create a single-page app in a single HTML file with the following requirements: - Name: Festival Lights Show - Goal: Control a virtual light show. - Features: Color changes, patterns, music sync. - The UI should be vibrant and dynamic.}
    \label{fig:qualitative_festival-lights-show}
    \end{subfigure}

    \caption{\textbf{Qualitative comparison of initialized UI \textit{vs.} refined UI.} Each row shows an initial UI (left), its revision based on function (middle), and its revision based on CUA's feedback (right).}
\vspace{-0.5cm}
\label{fig:qualitative}
\end{figure}

In Fig.\ref{fig:qualitative}, we present four representative revision cases—\textit{artisan-csa}, \textit{color-match-challenge}, \textit{csv-to-charts}, and \textit{festival-lights-show}. Each row displays the initial UI alongside its revised versions, evaluated under two criteria: Function Test and CUA Test.
Across the four cases, the revisions demonstrate distinct patterns of improvement. Revisions based on the Function Test, which addresses unsupported tasks, tend to focus on adding underlying functionality, sometimes with subtle visual changes. For example, the \textit{festival-lights-show} revision added a crucial ``Running'' state indicator, and the \textit{csv-to-charts} revision added a button to select a delimiter. In contrast, revisions based on the CUA Test consistently yield more significant visual modifications geared towards agent accessibility. For most websites, this meant adding buttons with clear boundaries and visual hints. In both \textit{color-match-challenge} and \textit{csv-to-charts}, both revision types improved accessibility by presenting more information and controls upfront, reducing the need for scrolling. A key CUA-friendly adaptation is seen in \textit{festival-lights-show}, where ``increase'' and ``reduce'' buttons were added as a complement to sliders, providing a more direct and reliable interaction method for agents.

%% file: section/conclusion.tex
\section{Conclusion}

We introduced \bench, a new benchmark for automatic GUI development (52 applications; 1560 tasks with programmatic checkers), and a Coder–CUA collaboration framework that recasts UI design as an agent-native loop, with the Coder as Designer and the CUA as Judge. Central to this loop is the CUA Dashboard, which compresses long agent navigation trajectories into compact, interpretable summaries that reliably convert raw interactions into actionable revision signals. Empirically, task solvability is foundational—readily improved by failure-driven functional summarization—whereas CUA navigation remains the primary bottleneck; feedback-driven redesigns (\eg~de-stylization, higher contrast, simplified layouts) consistently raise execution success and robustness, highlighting the value of designing \emph{for} agents rather than merely adapting human-centric interfaces.

%% file: section/appendix.tex
\clearpage
\appendix
\startcontents[app]    
\printcontents[app]{l}{1}


\section{Additional Results}
\label{appendix:exps}
\input{section/appendix_sec/appendix_exps}

\section{Prompts Usage}
In this section, we display the prompt used by individual rules.

\input{section/appendix_sec/task_proposer_prompt}

\input{section/appendix_sec/coder_prompt}

\input{section/appendix_sec/cua_policy_prompt}

\input{section/appendix_sec/commenter_prompt}

\section{Full Statistics and Examples}
In Tab.\ref{tab:stats:full}, we display the full statistics and corresponding examples.
\input{tex/stats}

\label{appendix:vis}

%% file: section/appendix_sec/appendix_exps.tex
\begin{figure*}[!h]
  \centering
  \setlength{\tabcolsep}{6pt}
  \renewcommand{\arraystretch}{1.0}

  \begin{minipage}{0.6\textwidth}
    \centering
    \captionsetup[subfigure]{aboveskip=0pt,skip=0pt}
    \subcaptionbox{Function completeness.\label{fig:operator-func}}{%
      \includegraphics[width=0.48\linewidth]{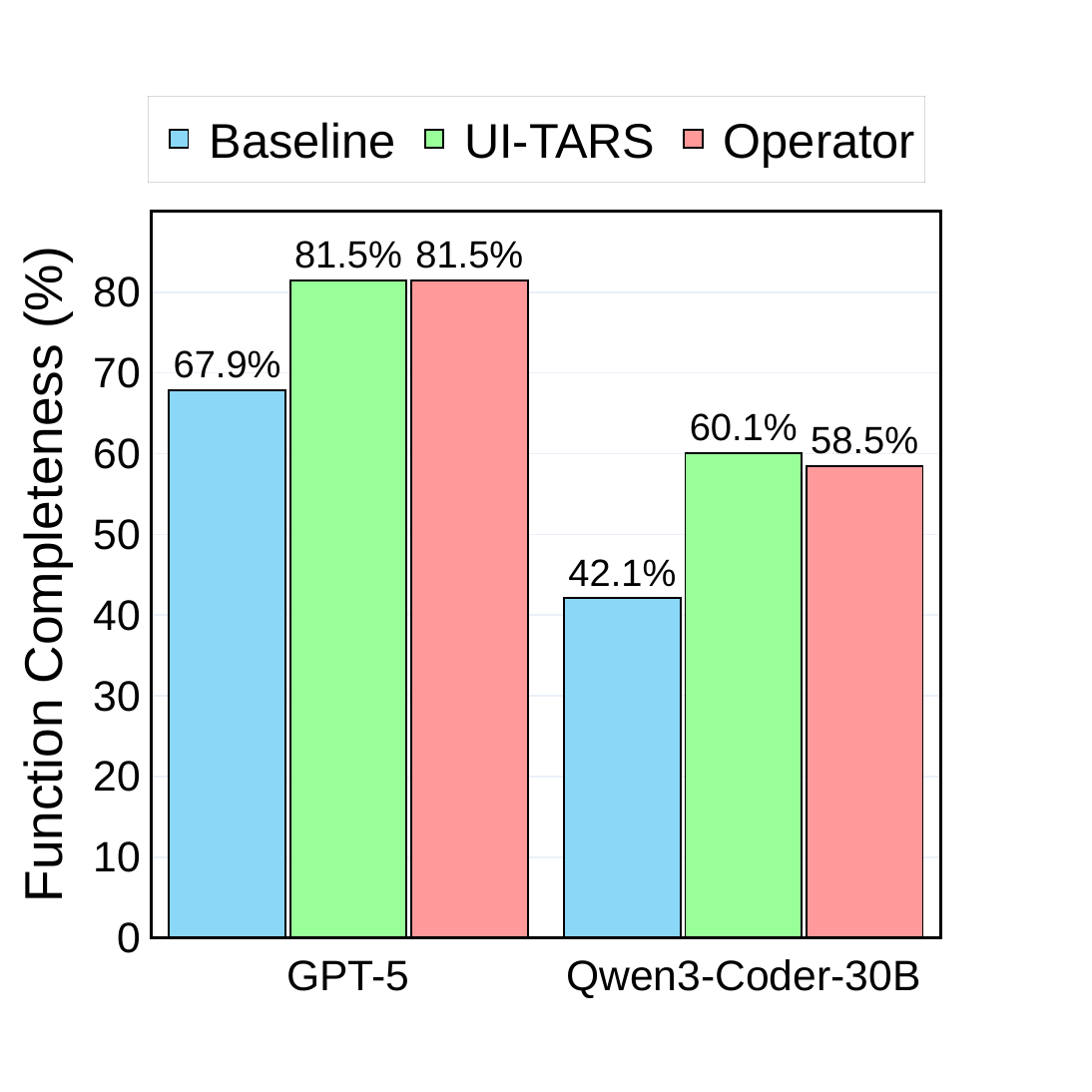}}
    \hfill
    \subcaptionbox{CUA success rate.\label{fig:operator-cua}}{%
      \includegraphics[width=0.48\linewidth]{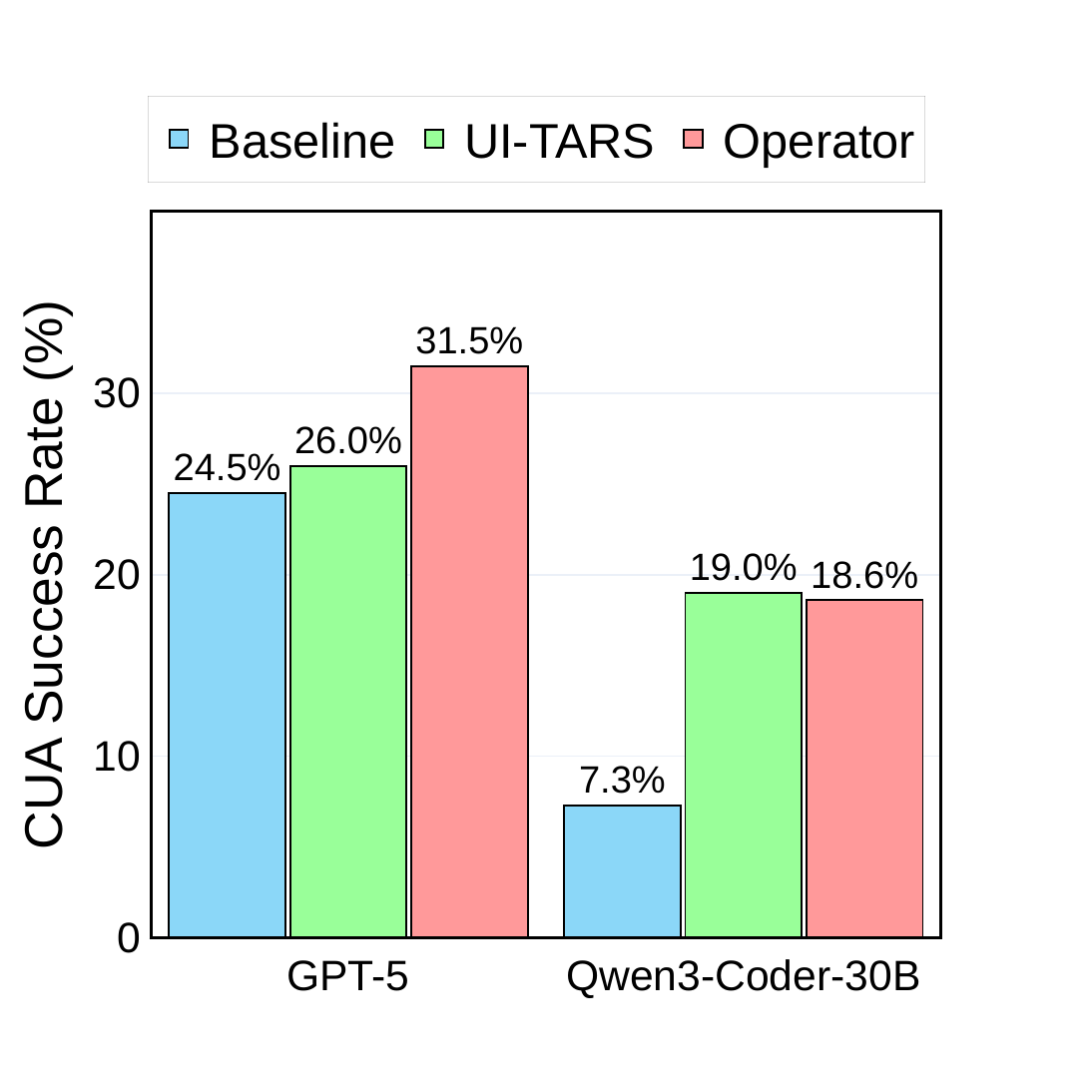}}
  \end{minipage}

  \caption{\textbf{Performance comparison after revision based on different CUA feedback.}}
  \label{fig:cua-choice}
\end{figure*}

\textbf{Effects by different CUAs choices.}
In Fig.~\ref{fig:cua-choice}, we compare UI-TARS and Operator as CUA policies within the integrated revision loop. We evaluate with two coders—GPT-5 (closed-source, stronger) and Qwen3-Coder-30B (open-source, weaker)—to cover both capability and licensing spectra. Both CUA policies yield comparable gains in functional completeness, with UI-TARS slightly outperforming on Qwen3-Coder-30B. Although the task-solvability signal is identical across CUAs, UI-TARS tends to fail more tasks, thereby surfacing richer failure cases and driving greater function-oriented revisions. For CUA success rate (SR), Operator delivers larger gains with the stronger coder (GPT-5), while improvements are similar across CUAs for the weaker coder. This suggests Operator’s navigation strengths are best realized on more complex UIs, whereas weaker coders often produce simpler interfaces. Overall, lightweight open-source CUAs like UI-TARS are an efficient and effective choice for harvesting navigation feedback in practice.

\begin{table*}[!h]
\centering
\caption{\textbf{Main results per model (Operator as CUA policy)}: Each cell shows A (B), where A represents the CUA Success Rate and (B) in parentheses denotes the Function Completeness Rate.}
\label{appendix:exps:operator}
\resizebox{\linewidth}{!}{%
\begin{tabular}{llccccccc}
\toprule
\textbf{Model} & \textbf{Version} & \textbf{landing} & \textbf{game} & \textbf{app} & \textbf{utility} & \textbf{interactive} & \textbf{tool} & \textbf{Overall} \\
\midrule
\multirow{2}{*}{GPT-5}
  & Baseline             & 34.7\% (53.0\%) & 24.8\% (77.8\%) & 27.3\% (70.6\%) & 14.4\% (63.3\%) & 18.1\% (73.0\%) & 21.9\% (70.0\%) & 24.5\% (67.9\%) \\
  & + Revise & 41.3\% (75.3\%) & 42.6\% (92.2\%) & 38.8\% (85.2\%) & 27.8\% (73.3\%) & 10.7\% (82.6\%) & 21.4\% (76.7\%) & \textbf{31.5\% (81.5\%)} \\
\midrule
\multirow{2}{*}{\makecell[l]{Qwen3-\\Coder-30B}}
  & Baseline             & 5.3\% (16.3\%)  & 9.3\% (50.4\%)  & 9.1\% (41.2\%)  & 11.7\% (43.9\%) & 7.0\% (52.2\%)  & 1.4\% (54.8\%)  & 7.3\% (42.1\%) \\
  & + Revise & 10.0\% (47.0\%) & 27.0\% (68.9\%) & 19.1\% (60.3\%) & 20.6\% (55.6\%) & 13.7\% (57.4\%) & 23.8\% (62.9\%) & \textbf{18.6\% (58.5\%)} \\
\midrule
\multirow{2}{*}{GPT-4o}
  & Baseline             & 4.7\% (9.7\%)   & 12.6\% (55.2\%) & 12.4\% (36.1\%) & 6.7\% (38.9\%)  & 9.3\% (44.8\%)  & 5.7\% (37.6\%)  & 8.8\% (36.3\%) \\
  & + Revise & 15.7\% (19.0\%) & 35.9\% (59.3\%) & 14.5\% (44.5\%) & 15.0\% (47.8\%) & 5.9\% (50.7\%)  & 13.8\% (46.2\%) & \textbf{16.9\% (43.8\%)} \\
\bottomrule
\end{tabular}%
}
\end{table*}

As shown in the Table~\ref{appendix:exps:operator}, when using operator as CUA policy for integrated revision, consistent improvements for both function completeness and CUA success rate can be observed. Moreover, compared to the CUA success rate showcased in Table~\ref{tab:unified_uitars_three_models_split}, it can be observed that Operator has higher CUA success rate than UI-TARS in hard domains such as game and app that requires responsive and complex interactions, showcasing its strong navigation capability.

\begin{table*}[!h]
\centering
\caption{\textbf{Dashboard Ablations}: CUA Success Rate (Function Completeness Rate), and token usage.}
\label{appendix:exps:commenter_ablation}
\resizebox{\linewidth}{!}{%
\begin{tabular}{lllllllllc}
\toprule
\textbf{Model} & \textbf{Dashboard} & \textbf{landing} & \textbf{game} & \textbf{app} & \textbf{utility} & \textbf{interactive} & \textbf{tool} & \textbf{Overall} & \textbf{Token (K)}  \\
\midrule
\multirow{3}{*}{GPT-5}
  & Text-only       & 24.0\% (50.7\%) & 31.1\% (87.8\%) & 21.2\% (69.4\%) & 16.1\% (55.6\%) & 8.9\% (59.3\%)  & 6.2\% (43.3\%)  & 18.7\% (62.1\%) & 3.2 \\
  & Screenshot-only & 17.3\% (30.3\%) & 16.7\% (65.6\%) & 12.4\% (42.7\%) & 15.6\% (38.3\%) & 5.2\% (27.8\%)  & 9.5\% (46.7\%)  & 12.8\% (41.7\%) & 15.5 \\
  & Dashboard       & 17.7\% (53.3\%) & 43.3\% (87.8\%) & 30.0\% (74.2\%) & 21.1\% (70.0\%) & 21.1\% (70.4\%) & 17.6\% (69.5\%) & \textbf{25.7\% (70.8\%)} & 4.3 \\
\midrule
\multirow{3}{*}{\makecell[l]{Qwen3-\\Coder-30B}}
  & Text-only       & 8.0\% (18.3\%)  & 20.7\% (61.9\%) & 7.3\% (42.4\%)  & 8.3\% (54.4\%)  & 10.7\% (48.9\%) & 16.2\% (57.1\%) & \textbf{11.7\%} (45.6\%) & 4.2 \\
  & Screenshot-only & 9.3\% (20.7\%)  & 11.9\% (63.7\%) & 5.2\% (34.5\%)  & 10.6\% (40.6\%) & 7.4\% (55.9\%)  & 5.2\% (37.6\%)  & 8.1\% (41.7\%) & 19.5 \\
  & Dashboard       & 6.7\% (23.3\%)  & 20.7\% (50.4\%) & 9.1\% (38.8\%)  & 11.1\% (49.4\%) & 12.2\% (39.3\%) & 11.4\% (55.2\%) & \textbf{11.7\%} (41.3\%) & 6.4 \\
\midrule
\multirow{3}{*}{GPT-4o}
  & Text-only       & 7.7\% (13.0\%)  & 14.8\% (57.0\%) & 12.7\% (34.8\%) & 2.8\% (37.8\%)  & 15.9\% (39.3\%) & 7.6\% (29.0\%)  & 10.8\% (34.8\%) & 2.8 \\
  & Screenshot-only & 4.7\% (10.3\%)  & 15.6\% (43.7\%) & 10.6\% (31.2\%) & 6.1\% (45.6\%)  & 5.6\% (34.8\%)  & 7.1\% (37.6\%)  & 8.5\% (32.5\%) & 14.8 \\
  & Dashboard       & 5.7\% (8.3\%)   & 31.5\% (55.2\%) & 10.0\% (28.2\%) & 8.3\% (34.4\%)  & 10.4\% (26.3\%) & 6.7\% (35.7\%)  & \textbf{12.3}\% (30.4\%) & 4.2 \\
\bottomrule
\end{tabular}%
}
\vspace{-1em}
\end{table*}

Table~\ref{appendix:exps:commenter_ablation} demonstrates the results when using different types of Dashboard for revision based on CUA navigation feedback. From the results, it can be inferred that dashboard is capable of providing comprehensive visual and textual cues derived from the CUA policy trajectories, but requiring the commenter to have strong visual perception. Moreover, our Dashboard yields a 70.4\% average token usage reduction across models and apps than screenshot-only commenter while delivering the strongest performance compared to
the variants, providing an efficient and effective way to utilize visual screenshots and textual histories to generate refinement insights.

\textbf{Why not use VLM-as-Judge as Verifiers.}
Table~\ref{appendix:exps:vlm_unreliable} demonstrates that why VLM-as-Judge evaluation on CUA task trajectory is unreliable. It can be observed that the VLM evaluation tends to judge the CUA policy outcome as failure compared to rule-based oracle evaluation, thus having very low balanced accuracy, recall and F1. Moreover, the low Cohen's $\kappa$ indicates very weak agreement of VLM evaluation compared to rule-based oracle evaluation. This indicates that VLM evaluation on the final screenshot only is unreliable, and may requires more screenshots along the CUA policy task trajectory for more reliable evaluation, leading to much higher computational cost.

\begin{table*}[!h]
\centering
\footnotesize
\caption{\textbf{VLM evaluation on final screenshot only is unreliable.} Given the final screenshot of CUA trajectory and the expected outcome, the accuracy of VLM evaluation is only slightly above the naive all-fail baseline; Balanced acc. is near 0.55; Recall/F1 and Cohen's $\kappa$ are low.}
\label{appendix:exps:vlm_unreliable}
{%
\begin{tabular}{lcccc}
\toprule
\textbf{Metric} & \textbf{Overall} & \textbf{GPT-5} & \textbf{Qwen2.5-VL-72B} & \textbf{GPT-4o} \\
\midrule
Naive all-fail baseline accuracy & 0.720 & 0.720 & 0.720 & 0.720 \\
Accuracy vs.\ oracle             & 0.735 & 0.736 & 0.738 & 0.732 \\
Balanced accuracy                & 0.556 & 0.549 & 0.568 & 0.552 \\
Precision (Pass)                 & 0.616 & 0.660 & 0.612 & 0.589 \\
Recall (Pass)                    & 0.147 & 0.121 & 0.178 & 0.142 \\
F1 (Pass)                        & 0.237 & 0.205 & 0.276 & 0.229 \\
Cohen's $\kappa$                 & 0.145 & 0.128 & 0.175 & 0.132 \\
\bottomrule
\end{tabular}%
}
\vspace{-1em}
\end{table*}

%% file: section/appendix_sec/task_proposer_prompt.tex
\begin{prompt}{Task Proposer Prompt}
\footnotesize
{
Propose a comprehensive set of 30 diverse, realistic user tasks for the following \{tag\_type\} application:\\
\\
Application: \{app\_title\}\\
Description: \{app\_description\}\\
\\
Each task should be:\\
- Clear and specific in its description\\
- Represent realistic user scenarios\\
- Cover different complexity levels and use cases\\
- Grounded in an observable outcome: The task's completion must be marked by a clear and unambiguous change in the application's state or interface. The expected outcome description must precisely define this terminal state.\\
- Avoid single element grounding (focus on complete workflows)\\
- Test the application's core functionality effectively\\
\\
\{tag\_specific\_content\}\\
\\
Tag Philosophy Template: \\
\\
Game:\\
Focus on GAME-SPECIFIC user tasks:\\
1. Playing complete game rounds or levels\\
2. Achieving high scores and personal bests\\
3. Completing specific game objectives or challenges\\
4. Using game controls and input methods\\
5. Navigating game menus and settings\\
6. Restarting games and trying different strategies\\
7. Progressing through difficulty levels\\
\\
Additional task requirements:\\
- Focus on actual gameplay actions and goals\\
- Include winning and losing scenarios\\
- Cover different skill levels and strategies\\
- Test game restart and replay functionality\\
- Emphasize user enjoyment and engagement\\
\\
Tool:\\
Focus on TOOL-SPECIFIC user tasks:\\
1. Creating or generating content using the tool\\
2. Inputting data in various formats and types (typed/pasted text or on-page controls)\\
3. Transforming and processing information\\
4. Previewing results in-page (no file uploads/downloads)\\
5. Using tool-specific features and options\\
6. Working with both simple and complex inputs\\
7. Completing end-to-end workflows within the page\\
\\
Additional task requirements:\\
- Focus on practical use cases and workflows\\
- Include both basic and advanced tool usage\\
- Cover different input types and scenarios without external files\\
- Verify visible in-page outputs or status changes in the DOM\\
- Emphasize real-world problem solving\\
\\
Utility:\\
Focus on UTILITY-SPECIFIC user tasks:\\
1. Setting up and configuring the utility for personal use\\
2. Adding, organizing, and managing data or items\\
3. Tracking progress and monitoring status over time\\
4. Using timers, reminders, and scheduling features\\
5. Customizing settings and preferences\\
6. Completing daily or routine activities\\
7. Accessing and updating information quickly\\
\\
Additional task requirements:\\
- Focus on everyday productivity scenarios\\
- Include setup and personalization tasks\\
- Cover routine and habitual usage patterns\\
- Test organization and tracking features\\
- Emphasize practical daily life applications\\
\\
Interactive:\\
Focus on INTERACTIVE-SPECIFIC user tasks:\\
1. Exploring and experimenting with interactive elements\\
2. Creating and manipulating visual or audio content\\
3. Adjusting parameters and settings in real-time\\
4. Playing with creative tools and features\\
5. Experiencing immersive visual or audio effects\\
6. Using touch, click, and gesture interactions\\
7. Customizing appearance and behavior\\
\\
Additional task requirements:\\
- Focus on creative and exploratory activities\\
- Include experimentation and play scenarios\\
- Cover different interaction methods\\
- Test customization and personalization\\
- Emphasize sensory and aesthetic experiences\\
\\
Landing:\\
Focus on LANDING-SPECIFIC user tasks:\\
1. Browsing and exploring page content and sections\\
2. Reading and understanding key information\\
3. Clicking on call-to-action buttons and links\\
4. Navigating through different page sections\\
5. Finding contact information and ways to engage\\
6. Viewing team, product, or service details\\
7. Accessing additional resources and links\\
\\
Additional task requirements:\\
- Focus on visitor browsing and exploration\\
- Include information-seeking behaviors\\
- Cover engagement and conversion actions\\
- Test navigation and content discovery\\
- Emphasize typical visitor journey scenarios\\
\\
App (default/other):\\
Focus on APP-SPECIFIC user tasks:\\
1. Creating, editing, and managing content or data\\
2. Using multiple features in combination\\
3. Setting up and personalizing the application\\
4. Completing complex multi-step workflows\\
5. Organizing and categorizing information\\
6. Accessing and updating saved information\\
\\
Additional task requirements:\\
- Focus on practical in-app usage\\
- Include multi-feature workflows and combinations\\
- Cover content creation and management\\
- Test personalization and customization\\
- Verify completion via visible state changes in the DOM (no external integrations)\\
\\
Task Categorization Framework:\\
\\
Each task must be categorized into one of the following three archetypes, which provides a structured approach to evaluating different facets of the application's functionality:\\
- "core\_function": Tests a single, primary feature in isolation.\\
- "user\_workflow": Tests a sequence of features that represent a complete user goal.\\
- "edge\_case": Tests non-standard inputs, boundary conditions, or less common interaction patterns.\\
\\
Please respond in JSON format:
\begin{verbatim}
{
  "app_name": "<app_name>",
  "tags": ["<tag1>", "..."],
  "tasks": [
    {
      "id": 1,
      "description": "Clear, specific task description",
      "category": "core_function|user_workflow|edge_case",
      "expected_outcome": "What should happen when task completes"
    }
  ]
}
\end{verbatim}
}
\end{prompt}

%% file: section/appendix_sec/coder_prompt.tex
\begin{prompt}{Coder Prompt}
\footnotesize
{
[Initial Website Generation]\\

Create a single-page web application based on the following specification:\\

\{instruction\}\\

Requirements:\\
1. Create a complete HTML file with embedded CSS and JavaScript\\
2. The app should be fully functional and interactive\\
3. Use modern HTML5, CSS3, and vanilla JavaScript (no external libraries)\\
4. Include proper semantic HTML structure\\
5. Make the UI clean, responsive, and user-friendly\\
6. Add unique IDs to interactive elements for easier automation testing\\
7. Ensure the app works in a 1280x720 viewport\\

Please generate the complete HTML file:\\

[Revision from CUA Failures — Core Prompt]\\

You are tasked with improving a web application based on detailed failure analysis from automated testing.\\

\#\# CONTEXT\\
Application: \{app\_name\}\\
Model: \{model\_name\}\\
Total Failed Tasks: \{len(failed\_tasks)\}\\
Failure Categories: \{list(failure\_categories.keys())\}\\
Original HTML Length: \{len(initial\_html.strip())\}\\

\#\# OUTPUT FORMAT\\
Generate a single, complete, and self-contained HTML file. The file must be fully functional, including all necessary CSS and JavaScript, from `!DOCTYPE html` to `/html`. Do not use placeholders or truncate the code.\\

\#\# ORIGINAL INITIAL WEBSITE (FULL)\\
\begin{lstlisting}
{initial_html}
\end{lstlisting}

\#\# COMMENTER UI ANALYSIS\\
\{(failure\_analysis or "No visual UI analysis available").strip()\}\\

\{(non\_regression\_contract\_prompt or '').strip()\}\\

\#\# IMPROVEMENT REQUIREMENTS\\

\#\#\# 1. Core Issues to Address\\
Based on the failure analysis, you must:\\
- Identify missing DOM elements that tasks expect to exist\\
- Add missing JavaScript functionality for user interactions\\
- Fix timing issues that prevent task completion\\
- Ensure proper event handling and state management\\
- Add missing visual feedback and UI updates\\

\#\#\# 2. Specific Fixes Needed\\
For each failed task category:\\
- **basic\_usage**: Ensure fundamental interactions work (clicking, displaying, updating)\\
- **workflow**: Support complete user workflows and multi-step processes\\
- **advanced\_feature**: Implement sophisticated UI behaviors and animations\\
- **edge\_case**: Handle unusual inputs and boundary conditions properly\\

\#\#\# 3. Technical Implementation Guidelines\\
- Preserve ALL existing working functionality from initial version\\
- Add missing HTML elements with unique IDs for automation\\
- Implement complete JavaScript event handlers and state updates\\
- Ensure synchronous UI updates for immediate feedback\\
- Do NOT introduce new input constraints that would block task inputs implied by the tasks (e.g., accept plain text or non-HTTP payloads if tasks need them). Validation must be permissive and never reduce what the initial version allowed.\\
- Do NOT auto-trigger flows on page load that would change initial states relied upon by tasks (e.g., auto-generation, auto-download, auto-navigation). Initial state should be neutral and idle.\\
- Keep critical controls visible within a 1280x720 viewport without scrolling. Avoid multi-panel "hub" layouts; prefer single-view, compact layouts that fit important controls on screen.\\
- Avoid adding non-essential animations/transitions; prioritize high visibility and clarity over decoration.\\
- Make sure timers, counters, and dynamic content work correctly\\

\#\#\# 4. DOM Structure Requirements\\
- Every interactive element MUST have a unique ID\\
- Form controls must have proper event listeners\\
- Dynamic content areas must update immediately on state changes\\
- Visual feedback must be implemented for all user actions\\

\#\#\# 5. JavaScript Functionality Requirements\\
- All user interactions mentioned in failed tasks must be fully implemented\\
- State changes must be reflected in the DOM immediately\\
- Event handlers must properly update all related UI elements\\
- Any game logic, scoring, timing must be complete and functional\\

Surgical Revision Policy\\
- Preserve existing IDs; do not rename or remove working elements from initial version.\\
- Avoid large rewrites. Patch only the functions, event handlers, and minimal markup necessary to satisfy the failed/unsupported tasks.\\
- Preserve working logic from initial version; do not regress features that already work.\\
- Reuse existing elements/IDs for state wherever possible; only add new IDs if strictly necessary to expose the state of new logic.\\
- Preserve initial version immediacy semantics. Do NOT introduce extra confirmation steps as prerequisites where initial version achieved completion via immediate interactions. Implement functional logic first, then expose proxies from the same code path; never update proxies without the underlying state change.\\

Commenter JSON (if provided)\\
- If the COMMENTER UI ANALYSIS is a JSON object, prioritize applying entries in `actionable\_changes` precisely.\\
- Keep changes surgical and bounded by those actionable suggestions; do not broaden scope beyond them.\\

\#\# OUTPUT REQUIREMENTS\\
Generate a COMPLETE, FULLY FUNCTIONAL HTML file that:\\
1. Addresses ALL failure points identified in the analysis\\
2. Maintains existing successful functionality from initial version\\
3. Implements missing features causing task failures\\
4. Provides proper DOM elements for automation testing\\
5. Ensures immediate UI feedback for all user actions\\

[Revision — Agent-centric Design Principles]\\

While improving functionality, apply the following design principles to optimize the UI for automated agents. The goal is functionality and testability, not human aesthetics.\\

\#\#\# A. Visual Clarity and Simplicity\\
- Use a simple color scheme (e.g., black text on a white background).\\
- Avoid decorative elements that do not serve a functional purpose, such as animations, gradients, or shadows.\\
- Establish a clear visual hierarchy using typography and spacing. Logically group related controls.\\

\#\#\# B. Robust Agent Interaction\\
- All interactive controls must be clearly labeled and sized appropriately to be easily and unambiguously targeted by automation tools.\\
- Support keyboard-based interaction for all core functionality. Navigable elements should have clear focus indicators.\\
- Prioritize immediate state updates upon interaction. Avoid complex, multi-step confirmation dialogs for actions where direct manipulation is sufficient.\\
- All critical functionality should be accessible within a standard 1280x720 viewport without requiring scrolling.\\

\#\#\# C. Transparent State Management\\
- The DOM must serve as a reliable, single source of truth for the application's state.\\
- Ensure that any significant state change (e.g., a result is generated, a calculation is complete) is clearly and synchronously reflected in the DOM. This can be achieved by updating element attributes, text content, or values.\\
- Interactive elements and state indicators must have unique and stable IDs to facilitate reliable testing and interaction.\\

\#\#\# D. Versatile Input Handling\\
- For continuous inputs (like sliders), provide alternative discrete control mechanisms (e.g., step buttons, direct text input). No interaction should rely solely on pointer-dragging.\\
- Input validation should be permissive and should not block inputs that an automated task might reasonably provide.\\
- Distinguish between actions that cause immediate, reversible state changes (e.g., selecting an option) and those that trigger irreversible, multi-step processes (e.g., submitting a form).\\

\#\#\# E. Behavior Preservation\\
- Simplifying the visual design must not alter the core interaction logic.\\
- Any user action that was immediate in initial version must remain immediate in the revised version.\\

Please generate the complete improved HTML file:\\

[Revision from Unsupported Tasks]\\

You are tasked with improving a web application to support additional tasks that are currently unsupported.\\

\#\# CONTEXT\\
Application: \{app\_name\}\\
Model: \{model\_name\}\\
Total Unsupported Tasks: \{len(unsupported\_tasks)\}\\
Original HTML Length: \{len(initial\_html.strip())\}\\

\#\# OUTPUT FORMAT\\
Generate a single, complete, and self-contained HTML file. The file must be fully functional, including all necessary CSS and JavaScript, from `!DOCTYPE html` to `/html`. Do not use placeholders or truncate the code.\\

\#\# ORIGINAL INITIAL WEBSITE (FULL)\\
\begin{lstlisting}
{initial_html}
\end{lstlisting}

\#\# UNSUPPORTED TASKS ANALYSIS\\
\{unsupported\_summary\}\\

\#\# CODE PRESERVATION CONTRACT (Non-Regression)\\
\{'' if ablate\_no\_contract else (non\_regression\_contract\_prompt or '').strip()\}\\

\#\# IMPROVEMENT REQUIREMENTS\\

\#\#\# 1. Task Support Issues to Address\\
Based on the unsupported task analysis, you must ADD missing functionality:\\
- Add missing DOM elements that tasks expect to exist\\
- Implement missing JavaScript functionality for user interactions\\
- Add missing form controls and input handling\\
- Implement missing display areas and visual feedback\\
- Add missing navigation and UI components\\

\#\#\# 2. Implementation Guidelines\\
- PRESERVE all existing working functionality from initial version\\
- ADD new HTML elements with unique IDs for automation\\
- IMPLEMENT complete JavaScript event handlers for new features\\
- ENSURE new UI elements are properly styled and visible\\
- DO NOT introduce new input constraints that would block task inputs implied by tasks; validation must be permissive and must not reduce what the initial version allowed.\\
- DO NOT auto-trigger flows on load that change initial states (no auto-generation, auto-download, auto-navigation). Start in a neutral, idle state.\\
- FIT critical controls within a 1280x720 viewport without scrolling. Avoid multi-panel hub layouts and unnecessary panels that push controls below the fold.\\
- IMPLEMENT missing workflows and user interaction patterns\\

\#\#\# 3. DOM Structure Requirements\\
- Every new interactive element MUST have a unique ID\\
- New form controls must have proper event listeners\\
- New content areas must update appropriately on state changes\\
- New visual feedback must be implemented for added interactions\\

\#\#\# 4. JavaScript Functionality Requirements\\
- All new user interactions mentioned in unsupported tasks must be fully implemented\\
- New state changes must be reflected in the DOM immediately\\
- New event handlers must properly update all related UI elements\\
- Any new game logic, scoring, timing must be complete and functional\\

\#\# OUTPUT REQUIREMENTS\\
Generate a complete and fully functional HTML file that:\\
1. Maintains all existing functionality from initial version.\\
2. Adds the missing functionality required to support the new tasks.\\
3. Implements all necessary DOM elements and JavaScript for task support.\\
4. Ensures all new features are robust and testable.\\

Commenter JSON (if provided)\\
- If upstream provides a commenter JSON analysis with `actionable\_changes`, follow those changes first, precisely and surgically.\\

Surgical Revision Policy\\
- Preserve existing IDs; do not rename or remove working elements from initial version.\\
- Avoid large rewrites. Patch only the functions, event handlers, and minimal markup necessary to satisfy the failed/unsupported tasks.\\
- Preserve working logic from initial version; do not regress features that already work.\\
- Reuse existing elements/IDs for state wherever possible; only add new IDs if strictly necessary to expose the state of new logic.\\
- Preserve initial version immediacy semantics. Do NOT introduce extra confirmation steps as prerequisites where initial version achieved completion via immediate interactions. Implement functional logic first, then expose proxies from the same code path; never update proxies without the underlying state change.\\

Please generate the complete improved HTML file:
}
\end{prompt}

%% file: section/appendix_sec/cua_policy_prompt.tex
\begin{prompt}{CUA Policy Prompt}
\footnotesize
{
You are a GUI agent. You are given a task and your action history, with screenshots. You need to perform the next action to complete the task.\\
\\
\#\# Output Format\\
\begin{verbatim}
Thought: ...
Action: ...
\end{verbatim}

\#\# Action Space\\
\\
click(point='x1 y1')\\
left\_double(point='x1 y1')\\
right\_single(point='x1 y1')\\
drag(start\_point='x1 y1', end\_point='x2 y2')\\
hotkey(key='ctrl c') \# Split keys with a space and use lowercase. Also, do not use more than 3 keys in one hotkey action.\\
type(content='xxx') \# Use escape characters \textbackslash', \textbackslash", and \textbackslash n in content part to ensure we can parse the content in normal python string format. If you want to submit your input, use \textbackslash n at the end of content. \\
scroll(point='x1 y1', direction='down or up or right or left') \# Show more information on the \texttt{direction} side.\\
wait() \# Sleep for 5 s and take a screenshot to check for any changes.\\
finished(content='xxx') \# Use escape characters \textbackslash', \textbackslash", and \textbackslash n in content part to ensure we can parse the content in normal python string format.\\
\\
\#\# Note\\
- Use \{language\} in \texttt{Thought} part.\\
- Write a small plan and finally summarize your next action (with its target element) in one sentence in \texttt{Thought} part.\\
\\
\#\# User Instruction\\
\{instruction\}
}
\end{prompt}

%% file: section/appendix_sec/commenter_prompt.tex
\begin{prompt}{Dashboard Commenter Prompt}
\footnotesize
{
You are diagnosing a UI design issue that caused a task failure for a Computer-Use Agent (CUA). Your goal is to conduct a root cause analysis based on a core set of design principles and output a structured diagnostic report in JSON format. This report will guide the next iteration of UI code generation.\\
\\
You will be provided with two images:\\
1. The current website state (Resolution: \{width\}x\{height\})\\
2. A storyboard summarizing the failed task attempt, arranged as a grid of step screenshots (variable count) fitted into a 1920x1080 canvas\\
\\
Your analysis must be guided by the following Agent-Centric UI Design Principles:\\
\\
1.  State Visibility: Any significant state change resulting from an agent's action must be clearly and synchronously reflected in the DOM. This can be achieved by updating element attributes, text content, or values. Ambiguous or out-of-band feedback (like temporary toast notifications) is considered a violation.\\
2.  Interaction Robustness: All UI components critical for task completion must be visible and actionable within a standard 1280x720 viewport without requiring scrolling. Elements should have clear, stable identifiers.\\
3.  Input Permissiveness: Input fields and controls should accept the most general data format required for the task, avoiding overly restrictive client-side validation that may block agent inputs.\\
4.  Predictable Behavior: The UI should remain in a stable, neutral state upon loading.\\
\\
Based on these principles, analyze the provided materials and output a compact JSON object.\\
\\
Output strictly as JSON with these keys only:\\
-   issues: An array of up to 3 short strings identifying the primary UI problem categories, derived from the violated principles (e.g., ``visibility'', ``interaction'', ``feedback'').\\
-   actionable\_changes: An array of 3--6 diagnostic statements. Each statement must identify a specific UI element (referencing selectors/IDs) and explain which design principle it violated, providing a root cause for the failure. Example: ``The element '\#submit-btn' violates the Interaction Robustness principle, as it is not visible in the default viewport.''\\
-   fit\_within\_screen: A diagnostic boolean flag. Set to \texttt{true} only if the primary reason for failure was a violation of the Interaction Robustness principle concerning viewport visibility.\\
-   avoid\_regressions: A confirmation flag, set to \texttt{true}, signifying that the diagnosis adheres to a "minimal intervention" philosophy. This confirms the analysis focuses solely on fixing the observed failure without disturbing unrelated, functional parts of the UI.\\
\\
Respond with JSON only, no extra text.\\
}
\end{prompt}

%% file: tex/stats.tex
\begin{table}[!t]
  \centering
  \scriptsize
\caption{\textbf{Distribution and examples of six domains in \bench.} For each domain, we show a website created by GPT-5, paired with 30 tasks (\textcolor{AuiTask}{blue}) simulating real-world usage. Each task is further linked to a rule-based verifier (\textcolor{AuiRule}{green}).}
  \setlength{\tabcolsep}{4pt}
  \renewcommand{\arraystretch}{1.12}
  \begingroup
  \begin{tabularx}{\textwidth}{lccXc}
    \toprule
    \textbf{Domain} & \textbf{\#Apps} & \makecell{\textbf{Percen-}\\\textbf{tage}} & \textbf{Example Instruction} & \textbf{GUI created by GPT-5} \\
    \midrule
    App & \makecell[t]{11} & \makecell[t]{21\%} &
    \parbox[t]{\linewidth}{\raggedright\scriptsize
      Create a single-page app in a single HTML file with the following requirements:\newline
      - Name: Healthy Meal Tracker\newline
      - Goal: Log meals and nutrition info.\newline
      - Features: Ingredient list, calories per meal, daily summary.\newline
      - The UI should be clean with food icons.\newline
      \textcolor{AuiTask}{\textbf{Task: Add five meals for today's date (any names/ingredients) so today's meal count reaches at least 5.}}\newline
      \textcolor{AuiRule}{\textbf{Rule: \texttt{\#dailyMealCount >= 5}}}
    } &
    \includegraphics[width=3.5cm, valign=t]{figures/app_screenshot.png} \\
    \midrule
    Landing & \makecell[t]{10} & \makecell[t]{19\%} &
    \parbox[t]{\linewidth}{\raggedright\scriptsize
      Create a single-page app in a single HTML file with the following requirements:\newline
      - Name: Nonprofit Impact Report\newline
      - Goal: Show measurable results of programs.\newline
      - Features: Infographics, success stories, donation link.\newline
      - The UI should be inspiring and visually engaging.\newline
      \textcolor{AuiTask}{\textbf{Task: Navigate to Success Stories and expand the first story card to reveal the full narrative.}}\newline
      \textcolor{AuiRule}{\textbf{Rule: \texttt{\#slides .slide:first-child button[aria-expanded] == 'true' OR \#slides .slide:first-child.expanded exists}}}
    } &
    \includegraphics[width=3.5cm, valign=t]{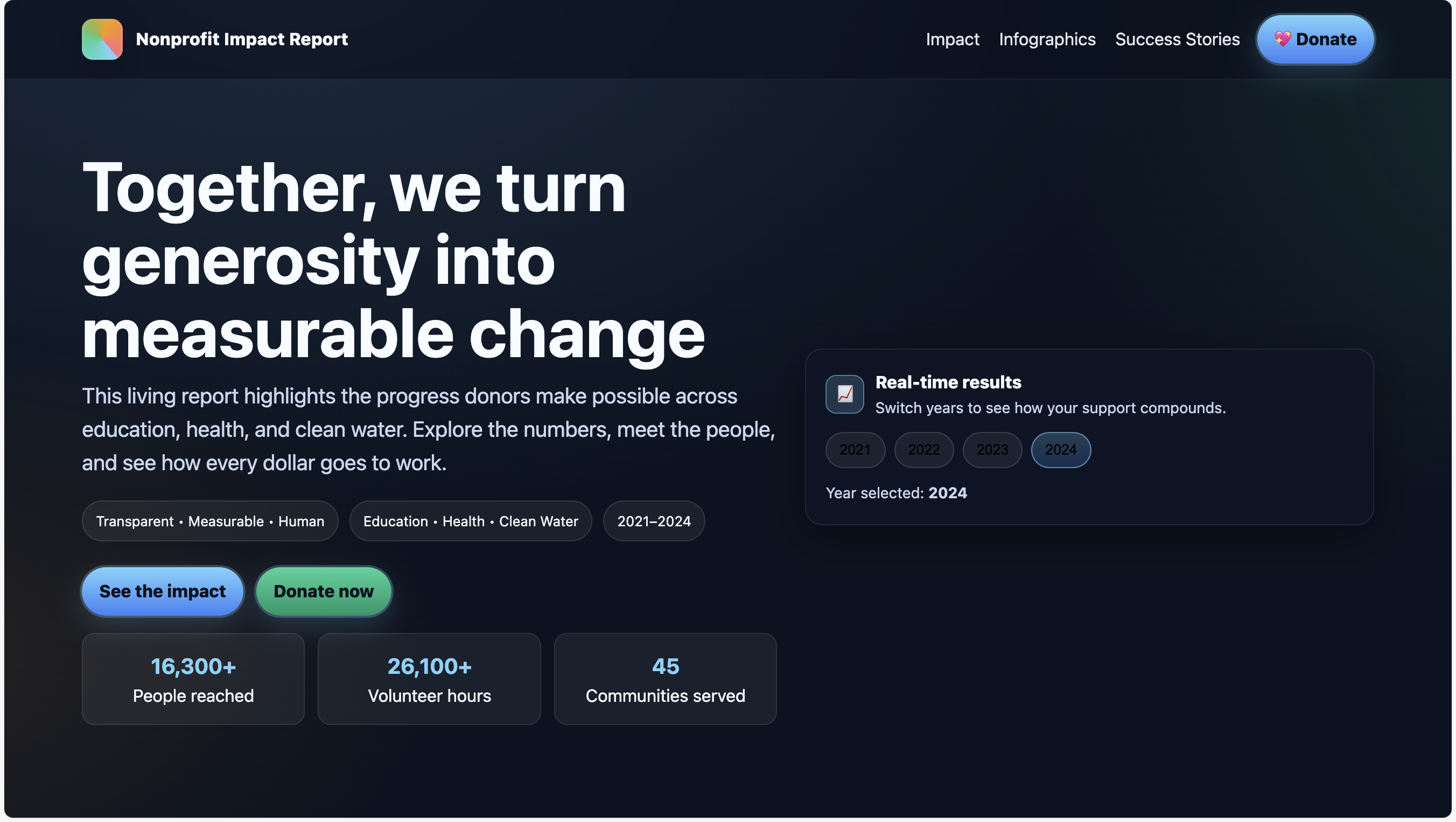} \\
    \midrule
    Game & \makecell[t]{9} & \makecell[t]{17\%} &
    \parbox[t]{\linewidth}{\raggedright\scriptsize
      Create a single-page app in a single HTML file with the following requirements:\newline
      - Name: Typing Rain\newline
      - Goal: Type falling words before they reach the bottom.\newline
      - Features: Increasing difficulty, accuracy tracker, score.\newline
      - The UI should be the city background with animated raindrop words.\newline
      \textcolor{AuiTask}{\textbf{Task: In a single run, achieve a score of at least 500 points.}}\newline
      \textcolor{AuiRule}{\textbf{Rule: \texttt{\#scoreValue >= 500}}}
    } &
    \includegraphics[width=3.5cm, valign=t]{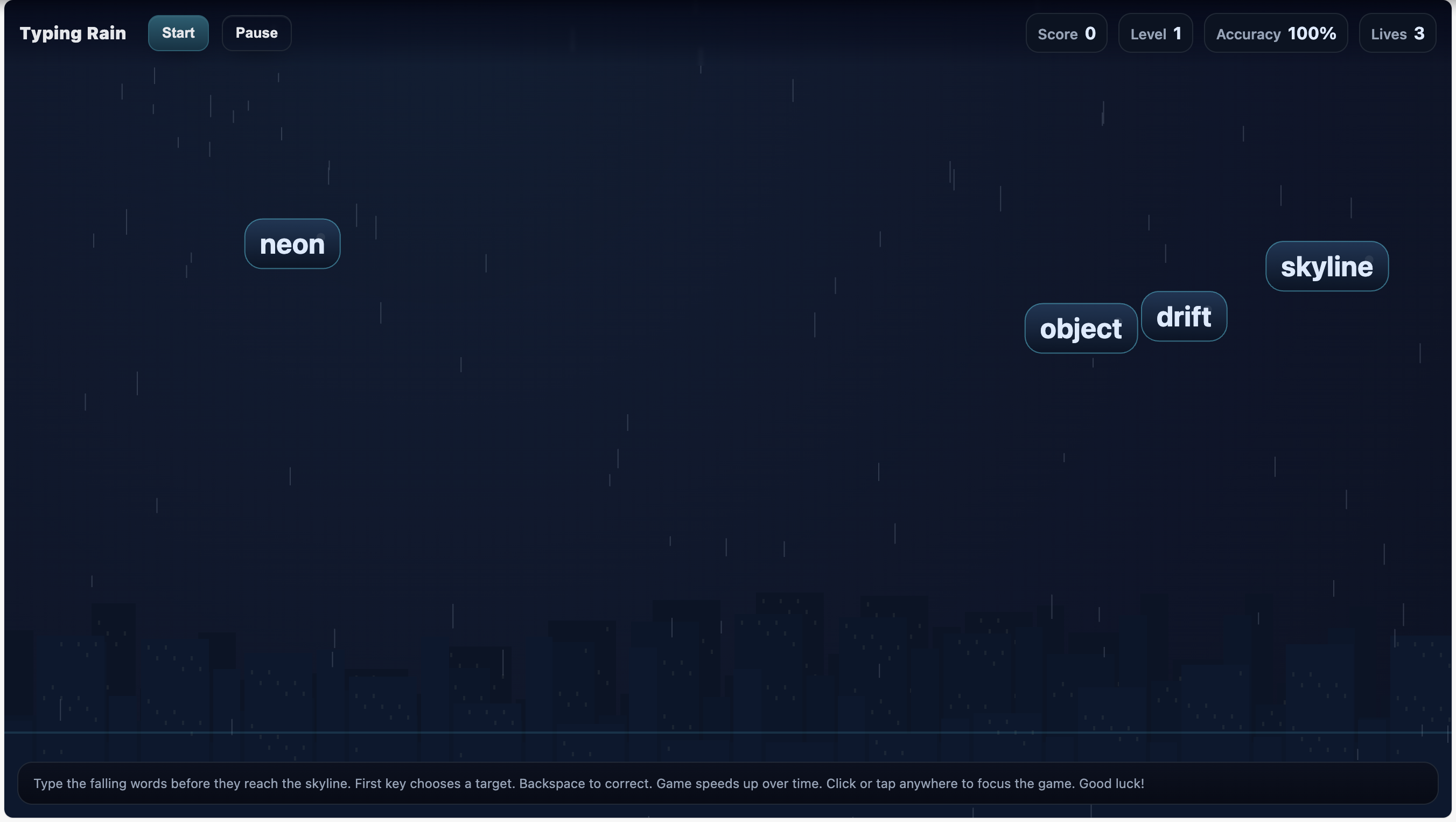} \\
    \midrule
    Interactive & \makecell[t]{9} & \makecell[t]{17\%} &
    \parbox[t]{\linewidth}{\raggedright\scriptsize
      Create a single-page app in a single HTML file with the following requirements:\newline
      - Name: Festival Lights Show\newline
      - Goal: Control a virtual light show.\newline
      - Features: Color changes, patterns, music sync.\newline
      - The UI should be vibrant and dynamic.\newline
      \textcolor{AuiTask}{\textbf{Task: Enable Music Sync, start playback, then pause the built-in track; confirm audio status is Paused while Music Sync remains enabled.}}\newline
      \textcolor{AuiRule}{\textbf{Rule: \texttt{\#audioStatus == 'Paused' AND \#syncBadge != 'Sync: Off'}}}
    } &
    \includegraphics[width=3.5cm, valign=t]{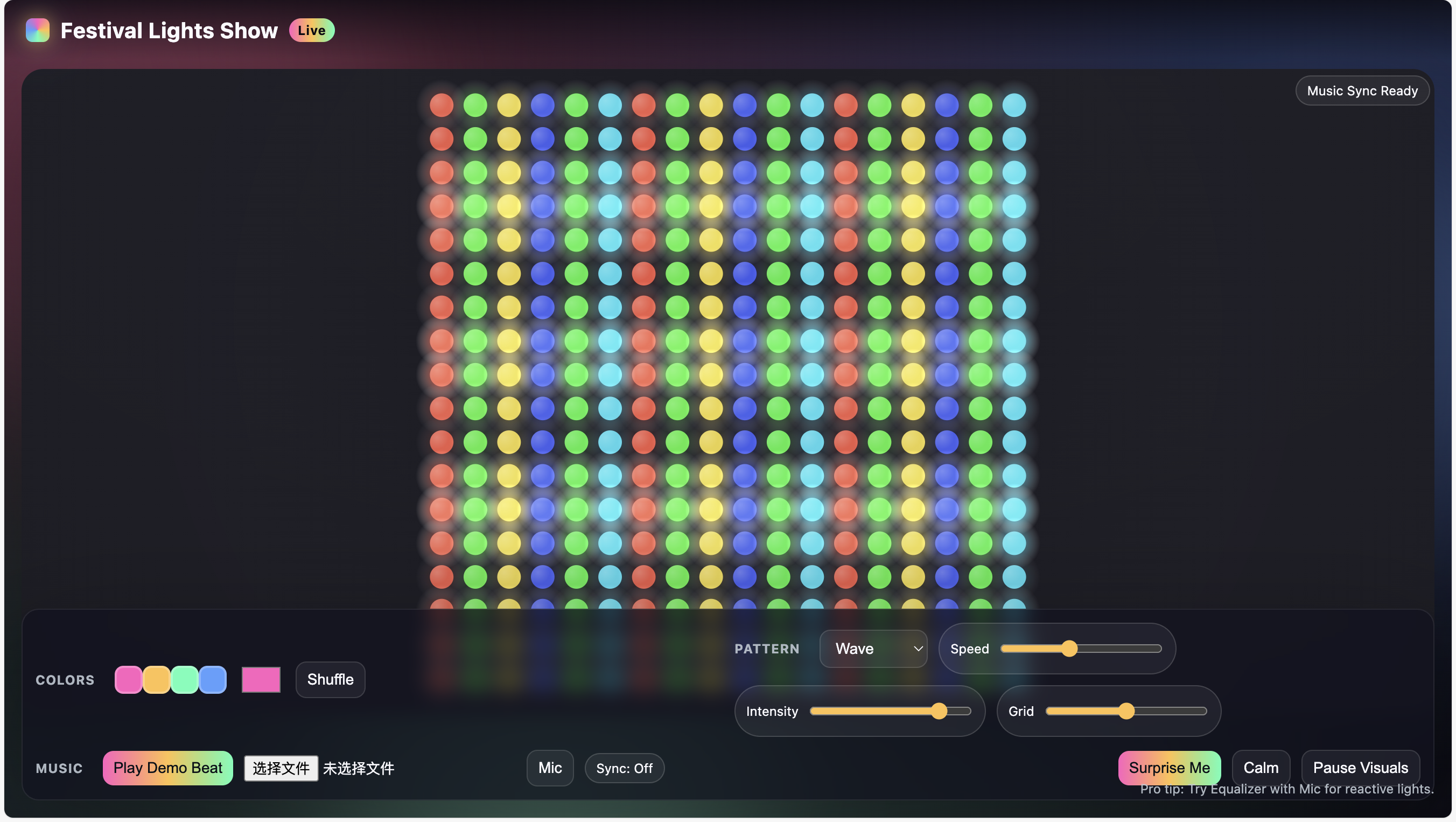} \\
    \midrule
    Tool & \makecell[t]{7} & \makecell[t]{13\%} &
    \parbox[t]{\linewidth}{\raggedright\scriptsize
      Create a single-page app in a single HTML file with the following requirements:\newline
      - Name: Customer Journey Flow\newline
      - Goal: Sketch customer journey stages and connections.\newline
      - Features: Add/edit stages, connect nodes, view JSON of the flow.\newline
      - The UI should be simple and full-screen.\newline
      \textcolor{AuiTask}{\textbf{Task: Create 'Social Ad' and 'Search Ad' leading to 'Landing Page', then to 'Consideration' and 'Purchase' (two branches merging into one path).}}\newline
      \textcolor{AuiRule}{\textbf{Rule: \texttt{\#io-json contains 'Social Ad' AND \#io-json contains 'Search Ad' AND \#io-json contains 'Landing Page' AND \#io-json contains 'Consideration' AND \#io-json contains 'Purchase'}}}
    } &
    \includegraphics[width=3.5cm, valign=t]{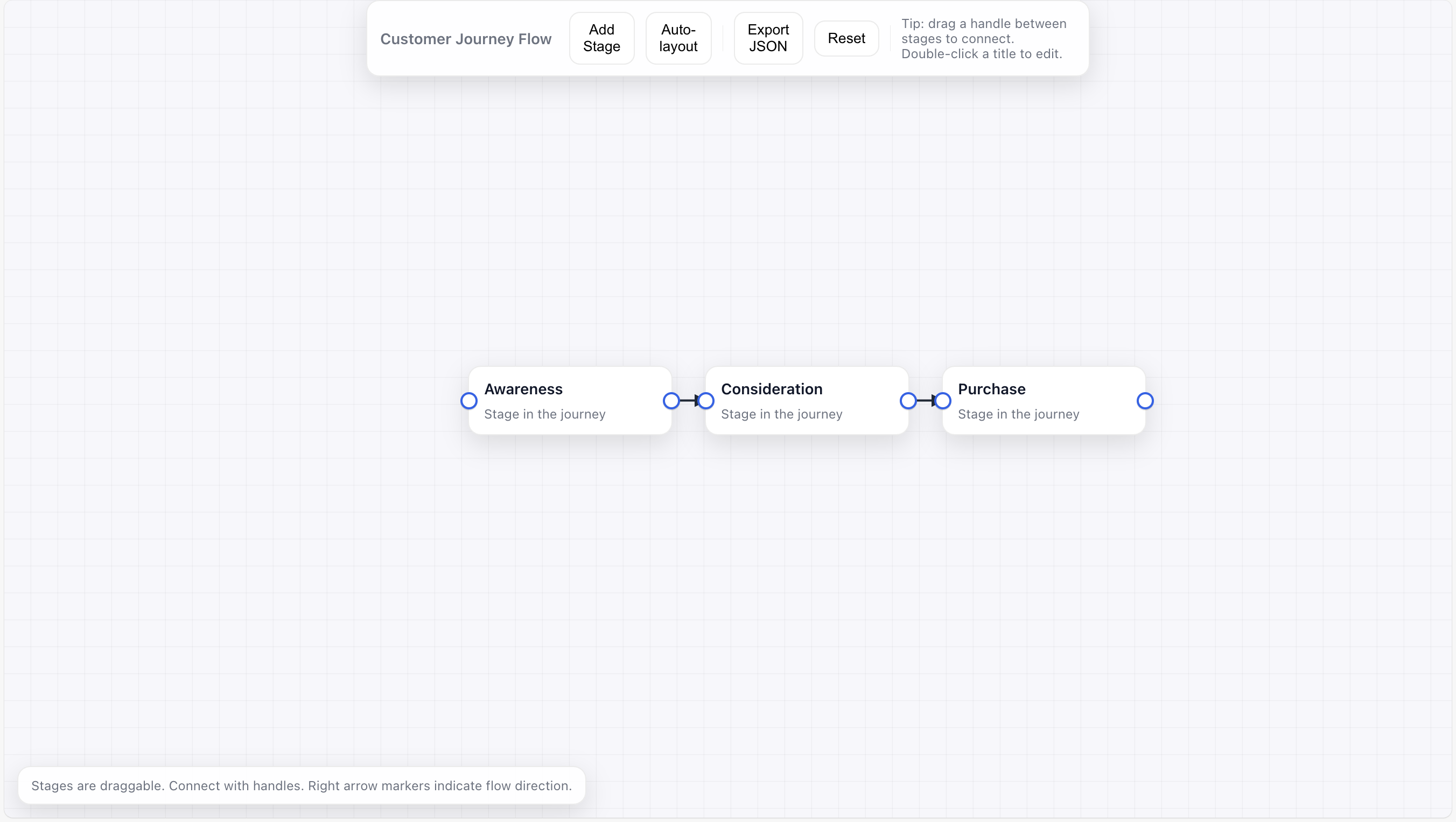} \\
    \midrule
    Utility & \makecell[t]{6} & \makecell[t]{12\%} &
    \parbox[t]{\linewidth}{\raggedright\scriptsize
      Create a single-page app in a single HTML file with the following requirements:\newline
      - Name: Pomodoro\newline
      - Goal: Time focus and break sessions.\newline
      - Features: Focus/break modes, timers, basic controls.\newline
      - The UI should be minimal and distraction-free.\newline
      \textcolor{AuiTask}{\textbf{Task: Start a short break and verify the mode label and starting time show a 5-minute break.}}\newline
      \textcolor{AuiRule}{\textbf{Rule: \texttt{\#lblSession == 'Short Break' AND \#lblTime contains '05:00'}}}
    } &
    \includegraphics[width=3.5cm, valign=t]{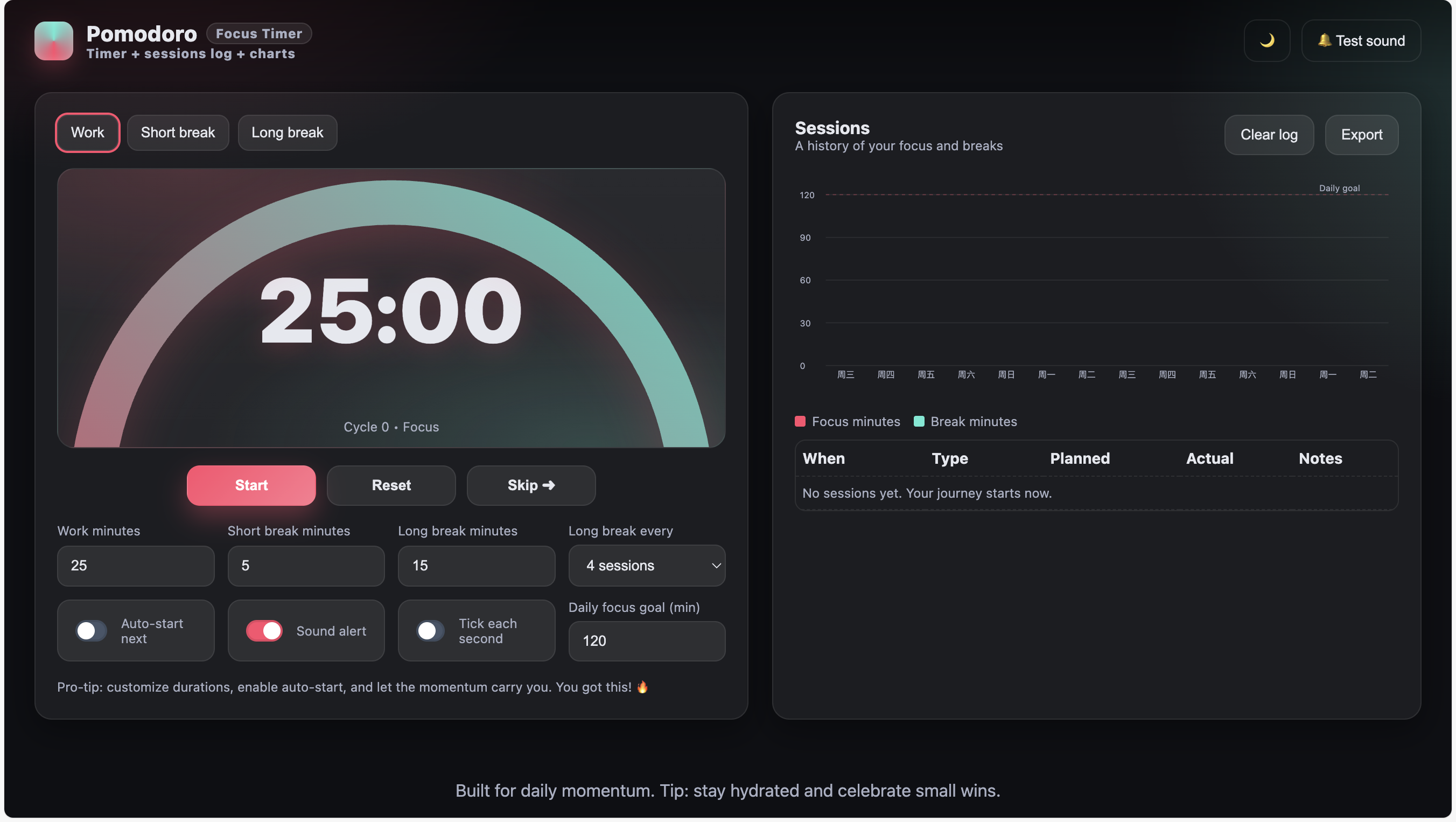} \\
    \bottomrule
  \end{tabularx}
  \endgroup
  \label{tab:stats:full}
\end{table}

%% file: main.bib
@article{toolformer,
  title={Toolformer: Language models can teach themselves to use tools},
  author={Schick, Timo and Dwivedi-Yu, Jane and Dess{\`\i}, Roberto and Raileanu, Roberta and Lomeli, Maria and Hambro, Eric and Zettlemoyer, Luke and Cancedda, Nicola and Scialom, Thomas},
  journal={Advances in Neural Information Processing Systems},
  volume={36},
  pages={68539--68551},
  year={2023}
}

@article{workarena,
  title={Workarena: How capable are web agents at solving common knowledge work tasks?},
  author={Drouin, Alexandre and Gasse, Maxime and Caccia, Massimo and Laradji, Issam H and Del Verme, Manuel and Marty, Tom and Boisvert, L{\'e}o and Thakkar, Megh and Cappart, Quentin and Vazquez, David and others},
  journal={arXiv preprint arXiv:2403.07718},
  year={2024}
}

@article{seeact,
  title={Gpt-4v (ision) is a generalist web agent, if grounded},
  author={Zheng, Boyuan and Gou, Boyu and Kil, Jihyung and Sun, Huan and Su, Yu},
  journal={arXiv preprint arXiv:2401.01614},
  year={2024}
}

@article{setofmask,
  title={Set-of-mark prompting unleashes extraordinary visual grounding in gpt-4v},
  author={Yang, Jianwei and Zhang, Hao and Li, Feng and Zou, Xueyan and Li, Chunyuan and Gao, Jianfeng},
  journal={arXiv preprint arXiv:2310.11441},
  year={2023}
}

@article{omniparser,
  title={Omniparser for pure vision based gui agent},
  author={Lu, Yadong and Yang, Jianwei and Shen, Yelong and Awadallah, Ahmed},
  journal={arXiv preprint arXiv:2408.00203},
  year={2024}
}

@inproceedings{showui,
  title={Showui: One vision-language-action model for gui visual agent},
  author={Lin, Kevin Qinghong and Li, Linjie and Gao, Difei and Yang, Zhengyuan and Wu, Shiwei and Bai, Zechen and Lei, Stan Weixian and Wang, Lijuan and Shou, Mike Zheng},
  booktitle={Proceedings of the Computer Vision and Pattern Recognition Conference},
  pages={19498--19508},
  year={2025}
}

@article{aguvis,
  title={Aguvis: Unified pure vision agents for autonomous gui interaction},
  author={Xu, Yiheng and Wang, Zekun and Wang, Junli and Lu, Dunjie and Xie, Tianbao and Saha, Amrita and Sahoo, Doyen and Yu, Tao and Xiong, Caiming},
  journal={arXiv preprint arXiv:2412.04454},
  year={2024}
}

@article{uground,
  title={Navigating the digital world as humans do: Universal visual grounding for gui agents},
  author={Gou, Boyu and Wang, Ruohan and Zheng, Boyuan and Xie, Yanan and Chang, Cheng and Shu, Yiheng and Sun, Huan and Su, Yu},
  journal={arXiv preprint arXiv:2410.05243},
  year={2024}
}

@article{videogui,
  title={VideoGUI: A Benchmark for GUI Automation from Instructional Videos},
  author={Lin, Kevin Qinghong and Li, Linjie and Gao, Difei and WU, Qinchen and Yan, Mingyi and Yang, Zhengyuan and Wang, Lijuan and Shou, Mike Zheng},
  journal={Advances in Neural Information Processing Systems},
  volume={37},
  pages={69329--69360},
  year={2024}
}

@inproceedings{uivision,
  title={UI-Vision: A Desktop-centric GUI Benchmark for Visual Perception and Interaction},
  author={Nayak, Shravan and Jian, Xiangru and Lin, Kevin Qinghong and Rodriguez, Juan A and Kalsi, Montek and Chapados, Nicolas and {\"O}zsu, M Tamer and Agrawal, Aishwarya and Vazquez, David and Pal, Christopher and others},
  booktitle={Forty-second International Conference on Machine Learning}
}

@misc{openaioperator,
  title={Operator}, 
  author={openai},
  year={2025},
  url={https://openai.com/index/introducing-operator/}, 
}

@article{claude,
  title={Claude 3.7 Sonnet System Card},
  author={Anthropic},
  year={2025}
}

@misc{uitars,
  title={UI-TARS-1.5}, 
  author={ByteDance Seed},
  year={2025},
  howpublished = {\url{https://seed-tars.com/1.5}}
}

@inproceedings{gaia,
  title={Gaia: a benchmark for general ai assistants},
  author={Mialon, Gr{\'e}goire and Fourrier, Cl{\'e}mentine and Wolf, Thomas and LeCun, Yann and Scialom, Thomas},
  booktitle={The Twelfth International Conference on Learning Representations},
  year={2023}
}

@inproceedings{alfred,
  title={Alfred: A benchmark for interpreting grounded instructions for everyday tasks},
  author={Shridhar, Mohit and Thomason, Jesse and Gordon, Daniel and Bisk, Yonatan and Han, Winson and Mottaghi, Roozbeh and Zettlemoyer, Luke and Fox, Dieter},
  booktitle={Proceedings of the IEEE/CVF conference on computer vision and pattern recognition},
  pages={10740--10749},
  year={2020}
}

@misc{habitat,
      title  = {Habitat 3.0: A Co-Habitat for Humans, Avatars and Robots},
      author = {Xavi Puig and Eric Undersander and Andrew Szot and Mikael Dallaire Cote and Ruslan Partsey and Jimmy Yang and Ruta Desai and Alexander William Clegg and Michal Hlavac and Tiffany Min and Theo Gervet and Vladimir Vondrus and Vincent-Pierre Berges and John Turner and Oleksandr Maksymets and Zsolt Kira and Mrinal Kalakrishnan and Jitendra Malik and Devendra Singh Chaplot and Unnat Jain and Dhruv Batra and Akshara Rai and Roozbeh Mottaghi},
      year={2023},
      archivePrefix={arXiv},
}

@inproceedings{minedojo,
  title     = {MineDojo: Building Open-Ended Embodied Agents with Internet-Scale Knowledge},
  author    = {Linxi Fan and Guanzhi Wang and Yunfan Jiang and Ajay Mandlekar and Yuncong Yang and Haoyi Zhu and Andrew Tang and De-An Huang and Yuke Zhu and Anima Anandkumar},
  booktitle = {Thirty-sixth Conference on Neural Information Processing Systems Datasets and Benchmarks Track},
  year      = {2022},
  url       = {https://openreview.net/forum?id=rc8o_j8I8PX}
}

@inproceedings{pix2code,
  title={pix2code: Generating code from a graphical user interface screenshot},
  author={Beltramelli, Tony},
  booktitle={Proceedings of the ACM SIGCHI symposium on engineering interactive computing systems},
  pages={1--6},
  year={2018}
}

@misc{unlocking,
    title={Unlocking the conversion of Web Screenshots into HTML Code with the WebSight Dataset},
    author={Hugo Laurençon and Léo Tronchon and Victor Sanh},
    year={2024},
    eprint={2403.09029},
    archivePrefix={arXiv},
    primaryClass={cs.HC}
}

@article{uilayout,
  title={UI Layout Generation with LLMs Guided by UI Grammar},
  author={Lu, Yuwen and Tong, Ziang and Zhao, Qinyi and Zhang, Chengzhi and Li, Toby Jia-Jun},
  journal={arXiv preprint arXiv:2310.15455},
  year={2023}
}

@inproceedings{adaptiveUIgrammar,
  title={Adaptive mobile interfaces through grammar induction},
  author={Kong, Jun and Ates, Keven L and Zhang, Kang and Gu, Yan},
  booktitle={2008 20th IEEE International Conference on Tools with Artificial Intelligence},
  volume={1},
  pages={133--140},
  year={2008},
  organization={IEEE}
}

@article{design2code,
  title={Design2code: Benchmarking multimodal code generation for automated front-end engineering},
  author={Si, Chenglei and Zhang, Yanzhe and Li, Ryan and Yang, Zhengyuan and Liu, Ruibo and Yang, Diyi},
  journal={arXiv preprint arXiv:2403.03163},
  year={2024}
}

@misc{webarena,
      title={WebArena: A Realistic Web Environment for Building Autonomous Agents}, 
      author={Shuyan Zhou and Frank F. Xu and Hao Zhu and Xuhui Zhou and Robert Lo and Abishek Sridhar and Xianyi Cheng and Tianyue Ou and Yonatan Bisk and Daniel Fried and Uri Alon and Graham Neubig},
      year={2024},
      eprint={2307.13854},
      archivePrefix={arXiv},
      primaryClass={cs.AI},
      url={https://arxiv.org/abs/2307.13854}, 
}

@article{gpt4o,
  title={Gpt-4o system card},
  author={Hurst, Aaron and Lerer, Adam and Goucher, Adam P and Perelman, Adam and Ramesh, Aditya and Clark, Aidan and Ostrow, AJ and Welihinda, Akila and Hayes, Alan and Radford, Alec and others},
  journal={arXiv preprint arXiv:2410.21276},
  year={2024}
}

@misc{gpt5blog,
  author       = {{OpenAI}},
  title        = {Introducing GPT-5},
  howpublished = {OpenAI website},
  year         = {2025},
  note         = {Available at \url{https://openai.com/index/introducing-gpt-5/}, accessed August 10, 2025},
}

@misc{qwen3coder,
  title={Qwen-3-coder}, 
  author={Qwen},
  year={2025},
  howpublished = {\url{https://qwenlm.github.io/blog/qwen3-coder}}
}

@misc{claude4intro,
  title={Introducing Claude 4}, 
  author={anthropic},
  year={2025},
  url={https://www.anthropic.com/news/claude-4}, 
}
